\documentclass{bmvc2k}

\title{Geometry-Grounded Unified 3D Perception for Autonomous Driving}

\addauthor{Longfei Xu}{xulongfei@buaa.edu.cn}{1,*}
\addauthor{Xiaohui Wang}{nekomio@bupt.edu.cn}{2,*}
\addauthor{Zehao Huang}{zehaohuang18@gmail.com}{}
\addauthor{Han Li}{lihan0620@buaa.edu.cn}{3}
\addauthor{Ya Yang}{yangya@bupt.edu.cn}{2}
\addauthor{Naiyan Wang}{winsty@gmail.com}{}
\addauthor{Si Liu}{liusi@buaa.edu.cn}{3}

\addinstitution{
 School of Computer Science and Engineering\\
 Beihang University\\
 Beijing, China
}
\addinstitution{
 School of Computer Science\\
 Beijing University of Posts and Telecommunications\\
 Beijing, China
}
\addinstitution{
 School of Artificial Intelligence\\
 Beihang University\\
 Beijing, China
}

\runninghead{Xu \bmvaEtAl}{Geometry-Grounded Unified 3D Perception}

\usepackage{booktabs}   
\usepackage{multirow}   
\usepackage{array}

\usepackage{amsmath}
\usepackage{amssymb}
\usepackage{amsfonts}
\usepackage{hyperref}

\usepackage{tikz}
\usepackage{graphicx}
\usepackage{amsmath}
\usepackage{xcolor}
\usetikzlibrary{arrows.meta,calc,positioning}

\makeatletter

\def\bmv@RenderAuthInstTwoColumn{
  \ifnum\bmv@ninsts>1
    \def\bmv@FN{FN}
  \else
    \def\bmv@FN{}
  \fi

  \hspace*{7mm}%
  \begin{minipage}[t]{0.40\textwidth}
    \def\bmv@action##1{%
      \bmv@unhbox{authname\bmv@FN##1}\\
      \bmv@unhbox{authmail##1}}
    \def\bmv@between{\\[3pt]}
    \expandafter\bmv@maplistaux\bmv@auths\bmv@endstop
  \end{minipage}
  \hfill%
  \begin{minipage}[t]{0.48\textwidth}
    \def\bmv@action##1{%
      \bmv@unhbox{inst\bmv@FN##1}}
    \def\bmv@between{\\[4pt]}
    \expandafter\bmv@maplistaux\bmv@insts\bmv@endstop
  \end{minipage}
}

\makeatother

\begin{document}

\maketitle

\begin{abstract}
Camera-based autonomous driving perception requires a shared representation that preserves metric 3D structure across synchronized multi-camera streams. However, existing image-based frameworks often rely on backbones pretrained for semantic recognition, and introduce 3D geometry through downstream task-specific modules. As a result, their shared representations may fail to preserve explicit metric geometry and consistent 3D scene structure. In this paper, we present a \textbf{Geo}metry-grounded \textbf{U}nified 3D \textbf{P}erception (\textbf{GeoUP}) framework that adapts the reconstruction-oriented latent of VGGT to calibrated, streaming multi-camera driving scenes. GeoUP factorizes cross-image interaction into self, temporal, and view attention to capture structurally distinct temporal and cross-view correspondences. It further injects calibration-aware raymap encodings to provide metric scale and camera geometry. The resulting geometry-grounded latent is decoded for metric depth estimation, 3D object detection, and semantic occupancy prediction, corresponding to surface-, instance-, and volume-level readouts of the same 3D scene. Through joint multi-task and multi-dataset training, GeoUP effectively leverages heterogeneous annotations and generalizes across diverse sensor configurations and perception ranges. Extensive experiments on nuScenes, Argoverse 2, Waymo, KITTI, and DDAD demonstrate that GeoUP achieves SOTA performance across detection, occupancy, depth estimation and end-to-end planning tasks. 
These results validate the effectiveness of geometry-grounded representations for unified 3D driving perception and planning.
Our project page is available at \url{https://buaa-colalab.github.io/geoup_page}.
\end{abstract}
\section{Introduction}
A central challenge in camera-based autonomous driving perception is to recover a metric and temporally consistent 3D scene representation from synchronized multi-view camera streams. Core 3D perception tasks can be regarded as different readouts of the underlying scene: depth estimation predicts image-aligned surface geometry~\cite{Monodepth2,DDAD}, 3D object detection localizes metric object instances~\cite{FCOS3D,DETR3D,BEVDet}, and semantic occupancy prediction estimates voxel-wise 3D scene states~\cite{TPVFormer,SurroundOcc,OccFormer}. Although these tasks differ in output format and granularity, they impose common requirements on the underlying representation, including semantic discriminability, metric geometry, multi-camera consistency, and temporal correspondence. This motivates a unified 3D perception representation~\cite{BEVFormer,PETRv2,BEVFusion,BEVFormerV2} that is organized around the 3D scene rather than being tailored to a specific perception task.

\begin{figure*}[t]
    \centering

    \subfigure[Recognition pretraining]{
        \label{fig:intro_recognition}
        \includegraphics[width=0.3\textwidth]{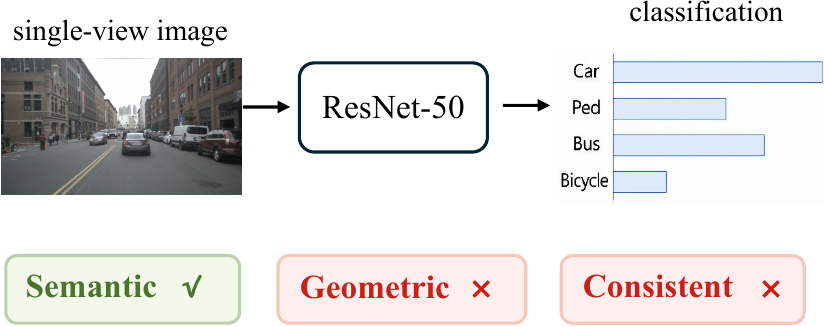}
    }
    \hfill
    \subfigure[Geometry pretraining]{
        \label{fig:intro_geometry}
        \includegraphics[width=0.3\textwidth]{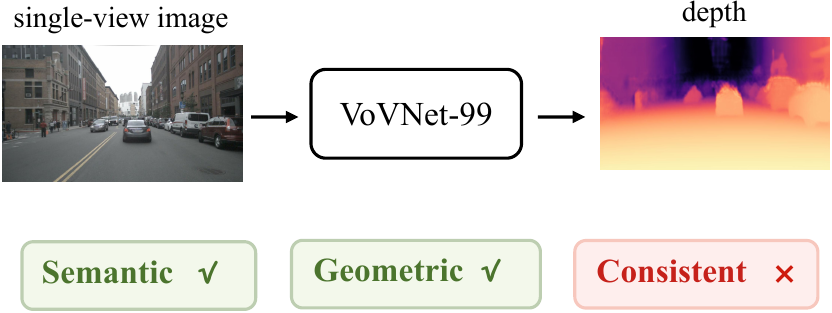}
    }
    \hfill
    \subfigure[3D reconstruction pretraining]{
        \label{fig:intro_reconstruction}
        \includegraphics[width=0.3\textwidth]{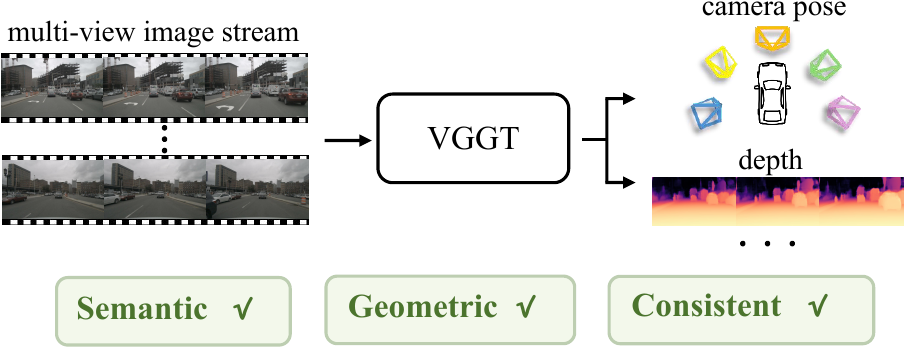}
    }

    \caption{
        \textbf{Comparison of pretraining paradigms for camera-based autonomous driving perception.}
        (a) Recognition pretraining, e.g., ImageNet~\cite{imagenet} classification with ResNet-50~\cite{ResNet}, learns semantic features but lacks geometry and multi-view consistency.
        (b) Geometry pretraining, e.g., monocular depth with VoVNet-99~\cite{vovnet-99}, adds geometric awareness but lacks cross-view and temporal modeling.
        (c) 3D reconstruction pretraining, e.g., VGGT~\cite{VGGT}, learns from multi-image geometric objectives, yielding a semantically rich, geometrically grounded, and temporally consistent representation.
    }
    \label{fig:intro}
\end{figure*}

The quality of such a unified representation depends critically on the geometric capacity of the underlying backbone. Most camera-based driving perception systems, however, still build their representations on general-purpose image backbones pretrained for objectives designed primarily for semantic recognition. These backbones, pretrained by supervised classification~\cite{ResNet,ViT}, self-supervised visual representation learning~\cite{DINOv2, MAE}, or vision-language alignment~\cite{CLIP}, provide semantically discriminative features, but do not explicitly encode metric 3D structure, as illustrated in Figure~\ref{fig:intro_recognition}. Consequently, geometry is often introduced by task-specific downstream modules, such as depth heads, BEV lifting~\cite{BEVDet,BEVFusion}, positional encodings~\cite{DETR3D,PETRv2}, or query-based geometric reasoning~\cite{PETRv2,BEVFormer}. 
Some methods further propose geometry-oriented pretraining, such as through monocular depth estimation ~\cite{vovnet-99, bevdepth} or monocular 3D detection~\cite{DD3D,FCOS3D,BEVFormerV2}, which improves geometric awareness but does not yield a multi-view, temporally consistent scene representation, as shown in Figure~\ref{fig:intro_geometry}. In contrast, visual geometry foundation models trained with multi-image 3D reconstruction objectives provide a more natural basis for learning such a representation, as shown in Figure~\ref{fig:intro_reconstruction}. This motivates us to explore whether such a shared representation can capture a coherent metric 3D scene and support heterogeneous perception tasks.




Among recent visual geometry foundation models~\cite{MapAnything,MASt3R,DUSt3R,pi3,streamvggt}, VGGT~\cite{VGGT} is a representative feed-forward model whose multi-image latent is structured by 3D geometry, while its DINOv2~\cite{DINOv2} image encoder provides semantically discriminative features. This combination makes VGGT an appealing starting point for autonomous driving perception.
However, vanilla VGGT~\cite{VGGT} is not directly suited to calibrated streaming multi-camera driving scenes. First, it does not explicitly inject camera calibration priors, leaving its representation unaware of the absolute metric scale that driving perception requires. 
Second, its global cross-image attention is computationally expensive and does not distinguish temporal from cross-view interactions in driving scenes.
These gaps motivate a driving-oriented adaptation with calibration-aware geometric encoding and structured spatiotemporal attention.

In this paper, we present a \textbf{Geo}metry-grounded \textbf{U}nified 3D \textbf{P}erception (\textbf{GeoUP}) framework that adapts VGGT~\cite{VGGT} to calibrated streaming multi-camera driving scenes. GeoUP retains the self-attention in VGGT~\cite{VGGT} and decomposes its global cross-image attention into temporal and view attention. Self-attention models image-level context, temporal attention captures correspondences within each camera stream, and view attention exchanges information across simultaneous cameras. To make the representation aware of camera geometry and absolute scale, GeoUP further injects patch-aligned raymap encodings~\cite{light, free3d} derived from camera intrinsics and global camera poses. With these adaptations, the shared latent representation learned by GeoUP supports metric depth estimation, 3D object detection, and semantic occupancy prediction, corresponding to surface-, instance-, and volume-level understanding of the same metric scene.
Furthermore, we introduce a joint multi-task and multi-dataset training strategy across these benchmarks, which strengthens the geometry-grounded representation and brings consistent gains. 

We evaluate GeoUP on widely used driving benchmarks, including nuScenes~\cite{nuScenes}, Argoverse 2~\cite{Argoverse2}, Waymo~\cite{Waymo}, KITTI~\cite{KITTI}, and DDAD~\cite{DDAD}, where it achieves state-of-the-art performance across detection, occupancy, and depth estimation.  Beyond perception, we further evaluate GeoUP as the visual backbone of an end-to-end planner. On the NAVSIMv2~\cite{navsim,Cao2025CORL} benchmark, GeoUP improves the EPDMS of DriveSuprim~\cite{drivesuprim} over the Depth Anything~\cite{yang2024depth} pretrained backbone. Together, these results demonstrate the effectiveness of adapting visual geometry foundation models for unified 3D driving perception and provide evidence that the learned representation transfers to downstream planning.

In summary, our contributions are:
\begin{itemize}
    \item We propose GeoUP, a geometry-grounded unified 3D perception framework that adapts the reconstruction-oriented latent of VGGT~\cite{VGGT} to autonomous driving and supports metric depth estimation, 3D object detection, and semantic occupancy prediction within a single architecture.
    \item We introduce driving-oriented adaptations for visual geometry foundation models, including temporal-view attention factorization and calibration-aware raymap injection, to handle streaming multi-camera inputs with structurally distinct temporal and cross-view interactions.
    \item We conduct joint multi-task and multi-dataset training across widely used autonomous driving datasets, showing that a geometry-grounded latent can benefit from heterogeneous supervision and generalize across surface-, instance-, and volume-level 3D perception tasks.
\end{itemize}

\section{Related Work}

\subsection{Camera-based Backbone Pretraining}

Camera-based 3D perception commonly relies on general-purpose image backbones, such as CNNs and vision transformers pretrained by supervised recognition~\cite{ResNet,ConvNeXt,ViT,SwinTransformer}, self-supervised learning~\cite{MoCo,SimCLR,DINO,DINOv2}, masked image modeling~\cite{BEiT,MAE,EVA,EVA02}, or vision-language supervision~\cite{CLIP,SigLIP}. These backbones provide strong semantic features, but their pretraining objectives mainly emphasize 2D appearance and semantic discrimination rather than metric 3D structure. To improve geometric awareness, some methods introduce depth, monocular 3D detection, or geometry-aware pretraining~\cite{DD3D,FCOS3D,BEVFormerV2}. However, these strategies are often single-frame or task-specific, and do not explicitly form a multi-view, temporal, metric scene representation. In contrast, our method adapts a reconstruction-oriented visual geometry backbone to calibrated streaming driving scenes, making geometry part of the shared representation.

\subsection{Task-specific Geometry Modeling}

Metric geometry is widely used in camera-based depth estimation, 3D detection, and occupancy prediction. Depth estimation directly predicts image-aligned surface geometry~\cite{Monodepth2,DDAD}; 3D detection introduces depth-aware lifting~\cite{BEVDet,LSS}, 3D query projection~\cite{DETR3D, mv2d, topo2d}, 3D positional encoding~\cite{PETR,PETRv2}, ray modeling~\cite{RayDN}, and temporal fusion~\cite{BEVFormer,StreamPETR} for metric object localization; occupancy prediction models voxel-wise or sparse point-wise scene states through volumetric, tri-plane, rendering-based, or query-based designs~\cite{TPVFormer,SurroundOcc,OccFormer, fb-occ, SparseOcc,OPUS}. Although effective, these methods usually model geometry inside task-specific pipelines. As a result, depth, detection, and occupancy may each learn separate geometric cues from a shared but weakly geometric image backbone. Our method instead treats them as complementary readouts from one geometry-grounded latent, covering surface-, instance-, and volume-level 3D perception.

\subsection{Visual Geometry Models}

Recent 3D reconstruction has shifted from specialized matching, structure from motion (SFM), and multi-view stereo (MVS) pipelines~\cite{COLMAP,SuperGlue,LoFTR,MVSNet,PatchmatchNet} toward visual geometry foundation models~\cite{DUSt3R,MASt3R,MUSt3R,VGGSfM,VGGT}. These models learn transferable geometric priors from reconstruction and correspondence objectives, and can predict cameras, depth, point maps, tracks, or dense correspondences from image inputs. VGGT~\cite{VGGT} is a representative feed-forward model that jointly predicts multiple geometric outputs with a reconstruction-oriented latent. However, such models are not directly tailored for autonomous driving. Driving data has calibrated multi-camera rigs, limited cross-view overlap, strong temporal continuity, and large metric-scale scenes. Recent VGGT-style~\cite{VGGT} driving adaptations focus mainly on reconstruction~\cite{dvgt,drivevggt}. Different from them, our method uses the visual geometry latent as a shared perception backbone and decodes it for depth estimation, 3D object detection, and semantic occupancy prediction.

\section{Method}

\begin{figure}[t]
    \centering
    \resizebox{\linewidth}{!}{
        \input{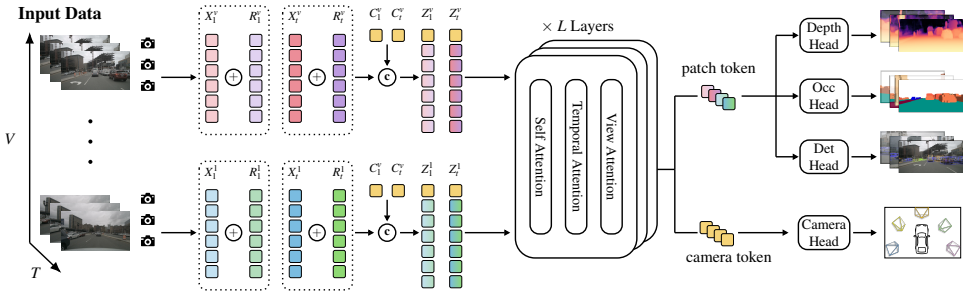}
    }
    \caption{
    \textbf{Overall pipeline of GeoUP.} Given streaming multi-view inputs, GeoUP constructs geometry-aware tokens $\mathbf{Z}$ by combining image patch tokens $\mathbf{X}$, raymap embeddings $\mathbf{R}$, and camera tokens $\mathbf{C}$. The tokens are processed by a factorized backbone with self, temporal, and view attention. The output patch tokens are used for depth estimation, 3D object detection, and occupancy prediction, while camera tokens are decoded for auxiliary camera pose prediction.
    }
    \label{fig:pipeline}
\end{figure}

We propose \textbf{GeoUP}, a unified perception framework that adapts VGGT~\cite{VGGT} to calibrated streaming multi-camera driving scenes. As shown in Figure~\ref{fig:pipeline}, GeoUP augments image patch tokens with calibration-derived raymap embeddings and camera tokens to form geometry-aware input representations, which are then processed by a spatiotemporal backbone with self, temporal, and view attention. The resulting token features are decoded by task-specific heads for metric depth estimation, 3D object detection, and semantic occupancy prediction. We describe the geometry-grounded backbone in Sec.~\ref{Backbone}, the perception heads in Sec.~\ref{Head}, and the training strategy in Sec.~\ref{Train}.

\subsection{Geometry-Grounded Backbone}
\label{Backbone}

GeoUP takes as input synchronized surround-view images from $V$ cameras over a temporal window of $T$ frames. We denote the image from camera $v$ at timestamp $t$ as $\mathbf{I}_{t}^{v}\in\mathbb{R}^{H\times W\times 3}$, where $v=1,\dots,V$ and $t=1,\dots,T$. The last timestamp $t=T$ is used as the current frame for downstream perception, and serves as the reference frame for each camera's temporal window. 
Each image is divided into patches with patch size $P$ and encoded by a DINOv2 encoder~\cite{DINOv2} into patch tokens $\mathbf{X}_{t}^{v}\in\mathbb{R}^{N\times C}$, where $N=HW/P^2$ and $C$ is the token dimension. To avoid redundant computation, GeoUP caches historical patch tokens and only encodes the current-frame images at each step. The full spatiotemporal token set is denoted as $\mathcal{X}=\{\mathbf{X}_{t}^{v}\in\mathbb{R}^{N\times C}\mid t=1,\dots,T, v=1,\dots,V\}$.

To inject explicit camera geometry, we encode each patch center as a Plücker ray~\cite{light, free3d} derived from the camera intrinsics and camera poses. For camera $v$ at time $t$, the raymap feature at grid location $(x,y)$ is computed as:
\begin{equation}
\mathbf{R}_{t}^{v}(x,y)
=
\mathrm{MLP}
\big(
\text{Proj}(x,y,\mathbf{K}_{t}^{v},\mathbf{E}_{t}^{v})
\big),
\end{equation}
where $\mathbf{K}_{t}^{v}\in\mathbb{R}^{4\times4}$ and $\mathbf{E}_{t}^{v}\in\mathbb{R}^{4\times4}$ denote the unnormalized camera intrinsic matrix and camera-to-reference poses (relative to camera $v$ at $t=T$), respectively. $\text{Proj}(\cdot)$ constructs the 6D Plücker representation from the patch center and camera parameters, followed by an MLP projects the result to the token dimension, yielding $\mathbf{R}_{t}^{v}\in\mathbb{R}^{N\times C}$. 
The geometry-enhanced token is then defined as $\tilde{\mathbf{X}}_{t}^{v}=\mathbf{X}_{t}^{v}+\mathbf{R}_{t}^{v}$.

\begin{figure}[t]
\centering

\resizebox{0.92\linewidth}{!}{
\input{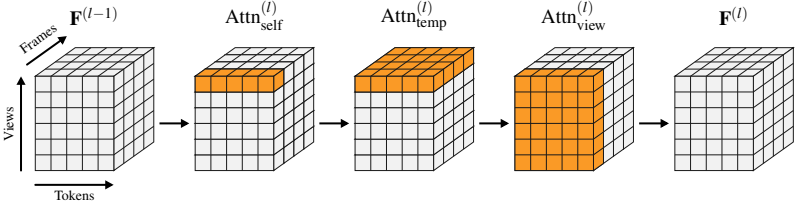}
}
\caption{
\textbf{Illustration of the factorized attention in GeoUP Transformer layers.} GeoUP applies self-attention within each image, temporal attention across frames from the same camera, and view attention across cameras at the same timestamp.
}
\label{fig:factorized_attn}
\end{figure}

Following VGGT~\cite{VGGT}, we prepend a camera token $\mathbf{C}_{t}^{v}\in\mathbb{R}^{1\times C}$ and register tokens $\mathbf{Q}_{t}^{v}\in\mathbb{R}^{4\times C}$ to each frame-view pair:
\begin{equation}
\mathbf{Z}_{t}^{v}
=
\text{Concat}( \left[
\mathbf{C}_{t}^{v};
\mathbf{Q}_{t}^{v};
\tilde{\mathbf{X}}_{t}^{v}
\right]).
\end{equation}
In VGGT~\cite{VGGT}, the first input image serves as the reference frame and uses a dedicated learnable camera token, while all other frames share a separate one. We adapt this to the multi-camera setting by treating the current frame ($t=T$) of each view as its reference frame. All reference frames share one learnable embedding, and all non-reference frames ($t < T$) share another.
The full set of tokens $\mathbf{Z} =\{\mathbf{Z}_{t}^{v}\mid t=1,\dots,T, v=1,\dots,V\}$ forms the initial backbone representation $\mathbf{F}^{(0)}$.
Then, $\mathbf{F}^{(0)}$ is refined through $L$ Transformer layers that alternate between intra-image and cross-image attention. The original VGGT~\cite{VGGT} design performs global cross-image attention over all input tokens, without distinguishing temporal and cross-view interactions. We instead factorize each layer into self-attention, temporal attention, and view attention. As illustrated in Figure~\ref{fig:factorized_attn}, the input features are organized along token, view, and frame dimensions, and each attention block operates on a specific grouping of this token volume. Let $l\in\{1,\dots,L\}$ denote the index of a backbone layer. The layer-wise update is formulated as:
\begin{equation}
\mathbf{F}^{(l)}
=
\text{Attn}_{\mathrm{view}}^{(l)}
\Big(
\text{Attn}_{\mathrm{temp}}^{(l)}
\big(
\text{Attn}_{\mathrm{self}}^{(l)}
(\mathbf{F}^{(l-1)})
\big)
\Big),
\end{equation}
where $\text{Attn}_{\mathrm{self}}$ operates within each image, $\text{Attn}_{\mathrm{temp}}$ aggregates tokens from the same camera across frames, and $\text{Attn}_{\mathrm{view}}$ exchanges information among cameras within each timestamp. This structured attention captures temporal continuity and cross-view geometric correspondence while avoiding the cost of full global attention.

After $L$ layers, the backbone produces per-layer token features $\mathbf{F}_{t}^{(l),v}$ for each timestamp, view and layer. We extract the image patch tokens from each layer as $\bar{\mathbf{X}}_{t}^{(l),v}$ for downstream perception tasks, discarding the camera and register tokens. Each task $a\in\{\mathrm{dep},\mathrm{det},\mathrm{occ}\}$ selects features from a task-specific subset of layer indices $\mathcal{I}_{a}$.

\subsection{Multi-task Perception Heads}
\label{Head}

Given the shared GeoUP features, we attach task-specific readout heads for depth estimation, 3D object detection, occupancy prediction, and camera pose prediction. These heads mainly follow existing task designs, while GeoUP focuses on providing a unified geometry-grounded representation for different levels of perception.

\noindent \textbf{Depth Head.}
For depth estimation, we use a DPT-based~\cite{DPT} dense prediction head to decode per-view metric depth from the selected current-frame GeoUP features $\{\bar{\mathbf{X}}_{T}^{(l),v}\}_{l\in\mathcal{I}_{\mathrm{dep}}}$. For each layer $l\in\mathcal{I}_{\mathrm{dep}}$, we inject camera intrinsics into the patch tokens before depth prediction:
\begin{equation}
\mathbf{Y}_{T}^{(l),d,v}
=
\bar{\mathbf{X}}_{T}^{(l),v}
+
\text{MLP}
\big(
\mathbf{K}_T^v
\big),
\end{equation}
where $\mathbf{K}_T^v\in\mathbb{R}^{4\times4}$ is the camera intrinsic matrix. This explicit conditioning provides camera geometry information to the depth branch and helps reduce the domain gap caused by different imaging geometries across datasets.

The depth head predicts a dense metric depth map $\mathbf{D}_T^v\in\mathbb{R}^{H\times W}$ for each current-frame view:
\begin{equation}
\mathbf{D}_T^v
=
\mathcal{H}_{\mathrm{dep}}
\big(
\{\mathbf{Y}_{T}^{(l),d,v}\}_{l\in\mathcal{I}_{\mathrm{dep}}}
\big).
\end{equation}
The prediction is scale-normalized metric depth, which is converted to real-scale depth using a predefined depth scale.

\noindent \textbf{Detection Head.}
For 3D object detection, we use a RayDN-based~\cite{RayDN} query detection head as the detection readout. The selected current-frame features ${\bar{\mathbf{X}}}_{T}^{(l),v}$ are passed through a detection neck to produce $\mathbf{Y}_T^{b,v}$.
Following Stream3DPPE~\cite{3DPPE}, we further inject depth-derived 3D position embeddings into the detection features. 
Based on the features, the detection head predicts 3D boxes, classification scores, and object velocities following RayDN~\cite{RayDN}. 

\noindent \textbf{Occupancy Head.}
For occupancy prediction, we use an OPUS-V2-style~\cite{OPUS} query decoder as the occupancy readout. Since occupancy requires scene-level spatial reasoning, this branch consumes multi-frame GeoUP features and performs temporal alignment inside the occupancy head. For each timestamp $t$, the selected multi-view and multi-level features are first processed by an occupancy neck to produce $\mathbf{Y}_{t}^{o,v}$.
The OPUS-V2-style decoder then takes the multi-frame and multi-view image features jointly and aggregates them through query-based feature sampling to predict semantic occupancy at the current timestamp:
\begin{equation}
\mathbf{O}_{T}
=
\mathcal{H}_{\mathrm{occ}}
\big(
\{\mathbf{Y}_{t}^{o,v}\}_{t=1,\dots,T;\,v=1,\dots,V}
\big),
\end{equation}
where $\mathbf{O}_{T}=\{(\mathbf{p}_{T,i},s_{T,i})\}_{i=1}^{N_{\mathrm{occ}}}$ denotes a set of sparse occupied points with semantic labels.

\noindent \textbf{Camera Head.}
Following VGGT~\cite{VGGT}, we keep the camera prediction branch during training. Although the camera poses are provided in input ray embedding, we find that explicit supervision helps this information flow through the features, as further validated by the ablation study in the appendix~\ref{camera}. It reads the camera-token features $\{\bar{\mathbf{C}}_{t}^{v}\}_{t=1,\dots,T,v=1,\dots,V}$ and predicts a 9D camera representation for each timestamp and view:
\begin{equation}
{{\mathbf{P}}}_{t}^{v}
=
\mathcal{H}_{\mathrm{cam}}
\big(
\{\bar{\mathbf{C}}_{t}^{v}\}_{t=1,\dots,T,v=1,\dots,V}
\big).
\end{equation}
This branch predicts camera-to-reference poses following the parametrization of VGGT~\cite{VGGT}, provides a camera-level training constraint together with the perception objectives.

\subsection{Training Strategy}
\label{Train}

We train GeoUP in two stages. In the first stage, we initialize the backbone from the pretrained VGGT~\cite{VGGT} weights and finetune it on driving datasets with the depth and camera branches:
\begin{equation}
\mathcal{L}_{\mathrm{ft}}
=
\mathcal{L}_{\mathrm{dep}}
+
\mathcal{L}_{\mathrm{cam}}.
\end{equation}
This stage adapts the reconstruction-oriented VGGT~\cite{VGGT} representation to driving scenes while preserving its geometric modeling capability.

In the second stage, we introduce the detection and occupancy heads and train GeoUP with a unified multi-task objective:
\begin{equation}
\mathcal{L}
=
m_{\mathrm{dep}}\lambda_{\mathrm{dep}}\mathcal{L}_{\mathrm{dep}}
+
m_{\mathrm{cam}}\lambda_{\mathrm{cam}}\mathcal{L}_{\mathrm{cam}}
+
m_{\mathrm{det}}\mathcal{L}_{\mathrm{det}}
+
m_{\mathrm{occ}}\mathcal{L}_{\mathrm{occ}},
\end{equation}
where $m_{\mathrm{dep}}$, $m_{\mathrm{cam}}$, $m_{\mathrm{det}}$, and $m_{\mathrm{occ}}$ indicate whether the corresponding annotation is available for the sampled data. This masking strategy enables joint training on multiple autonomous driving datasets with heterogeneous task annotations. Since the first-stage finetuning has already equipped GeoUP with strong depth and camera estimation ability, we set $\lambda_{\mathrm{dep}}=\lambda_{\mathrm{cam}}=0.1$ in the second stage to let the model focus more on the newly introduced detection and occupancy tasks while retaining geometry-oriented regularization.

The depth loss $\mathcal{L}_{\mathrm{dep}}$ combines metric depth regression with gradient-based regularization. The camera loss $\mathcal{L}_{\mathrm{cam}}$ supervises the normalized camera pose targets with L1 losses on translation, rotation, and intrinsic parameters, following VGGT~\cite{VGGT}. The detection and occupancy branches adopt the objectives of RayDN~\cite{RayDN} and OPUS-V2~\cite{OPUS}, respectively.


\section{Experiments}

\begin{table}[t]
\centering
\scalebox{0.63}{
\begin{tabular}{l|c|c|c|cc|ccccc} \hline
Method & Backbone & Backbone Init. & Input Size & mAP$\uparrow$ & NDS$\uparrow$ & mATE$\downarrow$ & mASE$\downarrow$ & mAOE$\downarrow$ & mAVE$\downarrow$ & mAAE$\downarrow$ \\ \hline
Stream3DPPE~\cite{3DPPE} 
& VoV-99 & FCOS3D 
& $800 \times 320$ & 50.0 & 58.5 & 0.565 & 0.261 & 0.376 & 0.251 & 0.200 \\

RoPETR~\cite{RoPETR} 
& VoV-99 & FCOS3D 
& $800 \times 320$ & 52.9 & 61.4 & 0.537 & 0.255 & 0.289 & 0.229 & \underline{0.195} \\

StreamPETR~\cite{StreamPETR} 
& EVA02-L & Cascade R-CNN
& $800 \times 320$ & 52.1 & 60.8 & 0.547 & \textbf{0.252} & 0.283 & 0.242 & 0.201 \\

RayDN~\cite{RayDN} 
& EVA02-L & Cascade R-CNN 
& $800 \times 320$ & 54.1 & 62.4 & 0.518 & \textbf{0.252} & 0.274 & 0.230 & \underline{0.195} \\ 

RayDN$\ddagger$~\cite{RayDN} 
& ViT-L & DINOv3 
& $800 \times 320$ & 49.9 & 55.3 & 0.599 & 0.256 & 0.290 & 0.638 & \textbf{0.184} \\ 

\textbf{GeoUP (Ours)} 
& ViT-L & VGGT 
& $800 \times 320$ & \underline{57.9} & \underline{64.4} & \underline{0.516} & 0.259 & \textbf{0.254} & \underline{0.223} & 0.204 \\ 

\textbf{GeoUP$^\dagger$ (Ours)} 
& ViT-L & VGGT 
& $800 \times 320$ & \textbf{59.2} & \textbf{65.3} & \textbf{0.496} & \underline{0.254} & \underline{0.271} & \textbf{0.217} & 0.196 \\
\hline
\end{tabular}
}
\caption{\textbf{Main results of 3D object detection on the nuScenes validation set~\cite{nuScenes}.}
GeoUP$^\dagger$ denotes the multi-dataset jointly trained model.
The Backbone Init.~column reports the source used to initialize the backbone.
RayDN$\ddagger$ denotes our reproduced variant trained with a ViT-L~\cite{dinov3} backbone using the official RayDN~\cite{RayDN} codebase.}
\label{tab:nuscenes_det_results}
\end{table}



\begin{table}[t]
\centering
\scalebox{0.75}{
\begin{tabular}{l|c|c|c|cc|ccc} \hline
Method & Backbone & Backbone Init. & Input Size & mAP$\uparrow$ & CDS$\uparrow$ & mATE$\downarrow$ & mASE$\downarrow$ & mAOE$\downarrow$ \\ \hline
StreamPETR~\cite{StreamPETR} & VoV-99 & FCOS3D & $960 \times 640$ & 20.3 & 14.6 & 0.843 & 0.321 & 0.650 \\
RayDN~\cite{RayDN} & VoV-99 & FCOS3D & $960 \times 640$ & 22.3 & 16.1 & 0.825 & 0.325 & 0.629 \\
Far3D~\cite{far3d} & VoV-99 & FCOS3D & $960 \times 640$ & 24.4 & 18.1 & 0.796 & \underline{0.304} & 0.538 \\
RayDN$\ddagger$~\cite{RayDN} & ViT-L & DINOv3 & $960 \times 640$ & 27.3 & 19.7 & 0.852 & 0.371 & 0.472 \\
Far3D~\cite{far3d} & ViT-L & Cascade R-CNN & $1536 \times 1536$ & 31.6 & 23.9 & \underline{0.732} & \textbf{0.303} & \underline{0.459} \\
\textbf{GeoUP (Ours)} & ViT-L & VGGT & $960 \times 640$ & \underline{37.2} & \underline{28.2} & 0.735 & 0.315 & 0.461 \\
\textbf{GeoUP$^\dagger$ (Ours)} & ViT-L & VGGT & $960 \times 640$ & \textbf{43.6} & \textbf{33.8} & \textbf{0.668} & \underline{0.304} & \textbf{0.428} \\ \hline
\end{tabular}
}
\caption{\textbf{Main results of 3D object detection on the Argoverse 2 validation set~\cite{Argoverse2}.}}
\label{tab:av2_det_results}
\end{table}

\subsection{Datasets and Metrics}

We evaluate 3D object detection on nuScenes~\cite{nuScenes},
Argoverse 2~\cite{Argoverse2}, and Waymo~\cite{Waymo}, following
their evaluation protocols. These benchmarks cover
camera configurations, perception ranges, and object category settings,
allowing us to evaluate the generalization of GeoUP across diverse
driving scenarios. We report mAP and NDS on nuScenes, mAP and CDS on
Argoverse 2, and LET-3D-APL, LET-3D-AP, and LET-3D-APH on Waymo.
Semantic occupancy prediction is evaluated on
Occ3D-nuScenes~\cite{occ3d}, which provides 3D semantic
annotations for surround-view scenes. We report mIoU and
RayIoU~\cite{SparseOcc}, including RayIoU under the 1,m, 2,m, and
4,m thresholds.
Depth estimation is evaluated on KITTI~\cite{KITTI} and
DDAD~\cite{DDAD} using Absolute Relative Error (Abs Rel) and the
threshold accuracy $\delta<1.25$.

\begin{table}[t]
\centering
\scalebox{0.75}{
\begin{tabular}{l|c|c|c|ccc} \hline
Method & Backbone & Backbone Init. & Input Size & mAPL$\uparrow$ & mAP$\uparrow$ & mAPH$\uparrow$ \\ \hline
PETRv2~\cite{PETRv2} 
& R101 & ImageNet 
& $1920 \times 1280$ & 36.6 & 51.9 & 47.9 \\

StreamPETR~\cite{StreamPETR} 
& R101 & ImageNet 
& $1920 \times 1280$ & 39.9 & 55.3 & 51.7 \\

MV-FCOS3D++~\cite{MV-Fcos3D++} 
& R101-DCN & FCOS3D++ 
& $1920 \times 1280$ & 38.5 & 53.2 & 50.0 \\

BEVFormer~\cite{BEVFormer} 
& R101-DCN & FCOS3D 
& $1920 \times 1280$ & 38.2 & 55.1 & 51.1 \\

DenseBEV++~\cite{DenseBEV} 
& R101-DCN & FCOS3D 
& $1920 \times 1280$ & 42.4 & 60.2 & 56.4 \\

RayDN$\ddagger$~\cite{RayDN} 
& ViT-L & DINOv3 
& $960 \times 640$ & 45.3 & 61.5 & 58.2 \\

\textbf{GeoUP (Ours)} 
& ViT-L & VGGT 
& $960 \times 640$ & \underline{51.5} & \underline{67.0} & \underline{63.5} \\

\textbf{GeoUP$^\dagger$ (Ours)} 
& ViT-L & VGGT 
& $960 \times 640$ & \textbf{54.3} & \textbf{70.7} & \textbf{67.7} \\ \hline
\end{tabular}
}
\caption{\textbf{Main results of 3D object detection on the Waymo validation set~\cite{Waymo}.}}
\label{tab:waymo_det_results}
\end{table}

\begin{table}[t]
\centering
\scalebox{0.65}{
\begin{tabular}{l|c|c|c|ccccc} \hline
Method & Backbone & Backbone Init. & Input Size & mIoU$\uparrow$ & RayIoU$\uparrow$ & RayIoU$_{1m}\uparrow$ & RayIoU$_{2m}\uparrow$ & RayIoU$_{4m}\uparrow$ \\ \hline
FB-Occ~\cite{fb-occ} (16f) 
& R50 & BEVDepth
& $704\times256$ & 39.1 & 33.5 & 26.7 & 34.1 & 39.7 \\

SparseOcc~\cite{SparseOcc} (16f) 
& R50 & Cascade R-CNN 
& $704\times256$ & 30.6 & 35.1 & 29.1 & 35.8 & 40.3 \\

OPUS-L~\cite{OPUS} (8f) 
& R50 & Cascade R-CNN 
& $704\times256$ & 36.2 & 41.2 & 34.7 & 42.1 & 46.7 \\

OPUS-V2-L~\cite{OPUS} (8f) 
& R50 & Cascade R-CNN 
& $704\times256$ & 38.6 & 44.0 & 38.0 & 45.0 & 49.2 \\

OPUS-V2-L$\ddagger$~\cite{OPUS} (8f) 
& ViT-L & DINOv3 
& $704\times256$ & 39.9 & 45.2 & 39.0 & 46.3 & 50.4 \\

\textbf{GeoUP (Ours) (8f)} 
& ViT-L & VGGT 
& $704\times256$ & \underline{41.5} & \underline{45.9} & \underline{39.3} & \underline{47.1} & \underline{51.3} \\ 

\textbf{GeoUP$^\dagger$ (Ours) (8f)} 
& ViT-L & VGGT 
& $704\times256$ & \textbf{42.3} & \textbf{47.0} & \textbf{40.7} & \textbf{48.1} & \textbf{52.2} \\ \hline
\end{tabular}
}
\caption{\textbf{Semantic occupancy prediction results on Occ3D-nuScenes~\cite{occ3d}.} \emph{8f} and \emph{16f} denote models fusing temporal information from 8 or 16 frames, respectively. OPUS-V2-L$\ddagger$ denotes our reproduced variant trained with a ViT-L backbone~\cite{DINOv2} using the official OPUS-V2~\cite{OPUS} codebase. }
\label{tab:occ_main}
\end{table}
\subsection{Implementation Details}

The original VGGT~\cite{VGGT} backbone contains 24 attention blocks, which incurs substantial computational cost. 
For efficiency, we construct a compact 12-block variant, denoted as VGGT-12, by selecting every other layer from the pretrained VGGT~\cite{VGGT} checkpoint.
Unless otherwise specified, the length of the temporal window is 4 frames. For perception tasks, we follow the default configurations of the corresponding heads. The RayDN-style~\cite{RayDN} detection head uses 900 object queries and retains queries from 4 frames. The OPUS-V2-style~\cite{OPUS} occupancy head uses 8-frame image features for sparse semantic occupancy prediction. The depth head decodes metric depth from the shared geometry-grounded representation.

We evaluate GeoUP under both single-dataset and multi-dataset training settings. For single-dataset training, we follow the schedules of prior methods for fair comparison. GeoUP is trained for 24 epochs on nuScenes~\cite{nuScenes}, 6 epochs on Argoverse 2~\cite{Argoverse2}, and 24 epochs on Waymo~\cite{Waymo}. We use the AdamW optimizer~\cite{AdamW} with a batch size of 16. The initial learning rate is set to $4\times10^{-4}$ with a cosine annealing schedule~\cite{cos}. For multi-dataset joint training, the dataset sampling ratio is designed according to the scale of each dataset. Argoverse 2~\cite{Argoverse2} is downsampled by a factor of 4, consistent with its shorter single-dataset training schedule, while Waymo~\cite{Waymo} is sampled every five frames. Accordingly, the sampling ratio for nuScenes~\cite{nuScenes}, Argoverse 2~\cite{Argoverse2}, Waymo~\cite{Waymo}, DDAD~\cite{DDAD}, and KITTI~\cite{KITTI} is set to $8:8:9:3:1$. Since joint training over heterogeneous datasets is more difficult to converge, we train the multi-dataset model with twice the training epochs of the single-dataset setting. The joint model is trained with a batch size of 64. We use the Muon optimizer~\cite{muon} with an initial learning rate of $6\times10^{-4}$ and the same cosine annealing schedule~\cite{cos}.


\begin{table}[t]
\centering
\scalebox{0.85}{
\begin{tabular}{l|cc|cc} \hline
\multirow{2}{*}{Method} & \multicolumn{2}{c|}{KITTI} & \multicolumn{2}{c}{DDAD} \\
& Abs Rel$\downarrow$ & $\delta<1.25$$\uparrow$ & Abs Rel$\downarrow$ & $\delta<1.25$$\uparrow$ \\ \hline
StreamVGGT~\cite{streamvggt} & \underline{0.102} & \underline{90.3} & 0.343 & 42.2 \\
VGGT~\cite{VGGT} & \underline{0.102} & 89.8 & 0.391 & 65.4 \\
MapAnything~\cite{MapAnything} & 0.210 & 52.2 & \underline{0.185} & \underline{77.1} \\
DVGT~\cite{dvgt} & 0.117 & 87.3 & 0.197 & 70.7 \\
\textbf{GeoUP (Ours)} & \textbf{0.072} & \textbf{93.8} & \textbf{0.114} & \textbf{89.5} \\
\textbf{GeoUP$^\dagger$ (Ours)} & 0.075 & 92.9 & 0.123 & 87.6 \\ \hline
\end{tabular}
}
\caption{\textbf{Depth estimation results on KITTI~\cite{KITTI} and DDAD~\cite{DDAD}.} }
\label{tab:depth_results}
\end{table}

\begin{table*}[t]
\centering
\scalebox{0.82}{
\begin{tabular}{l|ccccccccc|c} \hline
Backbone & NC$\uparrow$ & DAC$\uparrow$ & DDC$\uparrow$ & TL$\uparrow$ & EP$\uparrow$ & TTC$\uparrow$ & LK$\uparrow$ & HC$\uparrow$ & EC$\uparrow$ & EPDMS$\uparrow$ \\ \hline
DA-ViT-L~\cite{drivesuprim} & 98.4 & 98.6 & 99.6 & 99.8 & 90.5 & 97.8 & 97.0 & 98.3 & 78.6 & 87.1/90.5 \\
DINOv2-L~\cite{DINOv2} & 97.9 & 97.0 & 99.3 & 99.7 & 87.7 & 97.0 & 95.5 & 98.2 & 77.1 & 83.7/87.2 \\
DINOv3-L~\cite{dinov3} & 98.1 & 97.8 & 99.5 & 99.7 & 89.5 & 97.4 & 96.2 & 98.3 & 77.3 & 85.4/89.0 \\
\textbf{GeoUP (Ours)} & 98.4 & 98.8 & 99.5 & 99.9 & 91.9 & 98.0 & 97.4 & 98.3 & 78.8 & \textbf{87.9/91.4} \\ \hline
\end{tabular}
}
\caption{\textbf{Planning results on NAVSIMv2~\cite{navsim,Cao2025CORL}.} All variants use the same DriveSuprim~\cite{drivesuprim} planning decoder and training configuration, differing only in the visual backbone. DA-ViT-L denotes the Depth Anything~\cite{yang2024depth} backbone used in DriveSuprim.}
\label{tab:planning_results}
\end{table*}

\subsection{Main Results}

\noindent \textbf{3D Detection.}
Tables~\ref{tab:nuscenes_det_results}, \ref{tab:av2_det_results}, and \ref{tab:waymo_det_results} report the 3D object detection results on nuScenes, Argoverse 2, and Waymo, respectively.
GeoUP achieves strong performance across all three benchmarks, demonstrating the effectiveness and generality of the proposed geometry-grounded representation for camera-based 3D detection.
On nuScenes, GeoUP achieves 57.9\% mAP and 64.4\% NDS with a ViT-L backbone under the same input resolution as previous representative methods, outperforming RayDN by 3.8\% mAP and 2.0\% NDS.
We further reproduce RayDN with a DINOv3 ViT-L backbone, over which GeoUP achieves substantially better performance.
Interestingly, the EVA02-based RayDN also outperforms its DINOv3 counterpart, likely benefiting from its detection-oriented pretraining.
These comparisons further highlight the importance of geometric priors for 3D detection.
On Argoverse 2, GeoUP obtains 37.2\% mAP and 28.2\% CDS using an input size of $960 \times 640$, surpassing Far3D with a ViT-L backbone even though Far3D uses a much higher resolution of $1536 \times 1536$.
On Waymo, GeoUP also achieves clearly better results than previous camera-based detectors, reaching 51.5\% mAPL, 67.0\% mAP, and 63.5\% mAPH with a lower input resolution than prior methods.
We further report a multi-dataset jointly trained model, denoted as GeoUP$^\dagger$.
This model is trained on nuScenes, Argoverse 2, Waymo, DDAD, and KITTI, where each dataset only supervises the heads with available annotations.
Compared with the single-dataset model, GeoUP$^\dagger$ further improves the results on all three benchmarks, achieving 59.2\% mAP and 65.3\% NDS on nuScenes, 43.6\% mAP and 33.8\% CDS on Argoverse 2, and 70.7\% mAP and 67.7\% mAPH on Waymo.
These results indicate that heterogeneous supervision from datasets with diverse sensor configurations, scene distributions, and annotation types helps GeoUP learn a more robust geometry-grounded representation.

\noindent \textbf{Occupancy Prediction.}
Table~\ref{tab:occ_main} reports the semantic occupancy prediction results on the Occ3D-nuScenes benchmark. 
GeoUP achieves 41.5\% mIoU and 45.9\% RayIoU under single-dataset training, outperforming OPUS-V2-L with the same ViT-L scale and OPUS-V2-style decoder. This indicates that the improvement mainly comes from our geometry-grounded backbone. With multi-dataset joint training, our method further improves to 42.3\% mIoU and 47.0\% RayIoU, achieving the best results across all metrics.

\noindent \textbf{Depth Prediction.}
Table~\ref{tab:depth_results} reports the depth estimation results on KITTI and DDAD. GeoUP$^\dagger$ achieves strong depth prediction performance on both datasets. KITTI mainly consists of forward-facing stereo images with large view overlap, which is relatively favorable for visual geometry foundation models. Under this setting, VGGT and StreamVGGT already produce competitive depth estimates, while GeoUP$^\dagger$ further improves over VGGT by reducing Abs Rel from 0.102 to 0.075 and increasing $\delta<1.25$ from 89.8\% to 92.9\%. DDAD, in contrast, provides large-scale multi-camera driving scenes and better reflects the calibrated surround-view setting of autonomous driving. On this more challenging benchmark, generic visual geometry models such as VGGT and StreamVGGT show limited accuracy, while methods with explicit camera conditioning or driving-scene adaptation, such as MapAnything and DVGT, perform more favorably. GeoUP$^\dagger$ achieves the strongest results on DDAD, with 0.123 Abs Rel and 87.6\% $\delta<1.25$, showing the benefit of calibration-aware geometry modeling and unified training for metric depth prediction in multi-camera driving scenes.

\noindent \textbf{Transfer to End-to-End Planning.}
To assess whether GeoUP transfers beyond perception, we integrate it into DriveSuprim~\cite{drivesuprim} and evaluate it on the NAVSIMv2~\cite{navsim,Cao2025CORL} benchmark. We keep the planning decoder and training configuration unchanged and replace only the visual backbone, where the GeoUP features aggregate temporal information from four consecutive frames. To maintain comparability with the DriveSuprim results, we report EPDMS using both its original evaluation implementation and corrected NAVSIM evaluator, denoted as original/corrected.\footnote{\href{https://github.com/autonomousvision/navsim/issues/151}{NAVSIM Issue~\#151}}
As shown in Table~\ref{tab:planning_results}, DINOv2-L and DINOv3-L achieve 83.7\%/87.2\% and 85.4\%/89.0\% EPDMS, respectively, whereas DA-ViT-L (the Depth Anything~\cite{yang2024depth} pretrained ViT-L) used in DriveSuprim obtains 87.1\%/90.5\%. GeoUP reaches 87.9\%/91.4\%, outperforming DA-ViT-L by 0.8 and 0.9 points under the original and corrected evaluators, respectively. These results suggest that GeoUP provides a stronger representation for the DriveSuprim planner, supporting its transferability beyond perception tasks.

\begin{table}[t]
\centering
\scalebox{0.8}{
\begin{tabular}{cccc|cc|cc} \hline
\multirow{2}{*}{Global} & \multirow{2}{*}{View} & \multirow{2}{*}{Temporal} & \multirow{2}{*}{Pl\"ucker Ray} 
& \multicolumn{2}{c|}{Det.} & \multicolumn{2}{c}{Occ.} \\
& & & & mAP$\uparrow$ & NDS$\uparrow$ & mIoU$\uparrow$ & RayIoU$\uparrow$ \\ \hline
$\checkmark$ &  &  &  & 53.8 & 61.7 & 37.3 & 41.8 \\
 & $\checkmark$ &  &  & 53.3 & 61.1 & 37.5 & 41.4 \\
 &  & $\checkmark$ &  & 54.6 & 62.0 & 38.8 & 42.6 \\
 & $\checkmark$ & $\checkmark$ &  & \underline{55.5} & \underline{62.3} & \underline{39.9} & \underline{43.6} \\
 & $\checkmark$ & $\checkmark$ & $\checkmark$ & \textbf{56.4} & \textbf{63.0} & \textbf{40.3} & \textbf{44.5} \\ \hline
\end{tabular}
}
\caption{\textbf{Ablation study on the proposed backbone design} on the nuScenes validation set~\cite{nuScenes}. \emph{Global} denotes the original global cross-image attention in VGGT-12. \emph{View} and \emph{Temporal} denote the factorized cross-view and temporal attention designs, respectively. \emph{Pl\"ucker Ray} denotes the camera-aware geometric prior injected into the backbone.}
\label{tab:ablation_backbone_design}
\end{table}

\begin{table}[t] 
\centering 
\scalebox{0.80}{ 
\begin{tabular}{l|c|cc|cc|cc} \hline 
\multirow{2}{*}{Head Setting} & \multirow{2}{*}{Frames}
& \multicolumn{2}{c|}{Det.} & \multicolumn{2}{c|}{Occ.} & \multicolumn{2}{c}{Depth.}\\ & 
& mAP$\uparrow$ & NDS$\uparrow$ & mIoU$\uparrow$ & RayIoU$\uparrow$ & Abs Rel$\downarrow$ & $\delta<1.25$$\uparrow$ \\ \hline 
\multirow{4}{*}{Current-only heads} & 1 & 48.7 & 52.9 & 36.7 & 40.8 & \underline{0.102} & 90.1 \\ 
& 2 & 51.0 & 57.9 & 37.9 & 41.7 & \textbf{0.098} & \underline{90.6}\\ 
& 4 & \underline{52.8} & \underline{59.2} & \textbf{38.9} & \underline{42.2} & \textbf{0.098} & \textbf{90.7}\\ 
& 8 & \textbf{53.6} & \textbf{60.1} & \underline{38.7} & \textbf{43.3} & \textbf{0.098} & \textbf{90.7}\\ \hline 
\multirow{4}{*}{Temporal heads} & 1 & 55.1 & 62.5 & 40.8 & 45.0& 0.098 & 90.6 \\
& 2 & 55.7 & \underline{62.8} & 41.4 & 45.4 & \underline{0.094} & \underline{91.2}\\ 
& 4 & \underline{56.5} & \textbf{63.6} & \underline{41.5} & \underline{45.9} & \textbf{0.092} & \textbf{91.4}\\ 
& 8 & \textbf{56.9} & \textbf{63.6} & \textbf{41.6} & \textbf{46.1} & \textbf{0.092} & \textbf{91.4} \\ \hline 
\end{tabular} } 
\caption{\textbf{Ablation on the number of input frames} on the nuScenes validation set~\cite{nuScenes}. All models are trained for 50 epochs. \emph{Current-only heads} denote that the heads use only current-frame information, while \emph{Temporal heads} denote that the downstream heads also exploit temporal information.} \label{tab:frame_ablation} 
\end{table}

\subsection{Ablation Study}
\label{sec:ablation}

We conduct ablation studies mainly on the nuScenes validation set. Unless otherwise specified, all models are trained for 24 epochs with an input resolution of $672\times224$. 

\noindent \textbf{Effect of Driving-Oriented Backbone Adaptation.}
Table~\ref{tab:ablation_backbone_design} validates the proposed adaptations of the VGGT backbone to multi-camera driving scenes. We first compare cross-image attention strategies. In the global-attention setting, all images are jointly attended, following the original VGGT-style interaction, where the front camera of the current frame is used as the reference frame. For temporal attention, attention is only performed among images from the same view across different timestamps. To ensure that each temporal window has its own reference, we use the current frame of each view as the reference frame. Similarly, for view attention, we perform cross-view interaction within each timestamp and use the front camera of each frame as the reference frame.
Temporal attention achieves the best single-factorized performance, even surpassing global attention, thanks to temporal continuity within each camera stream. View attention is less effective, as the limited overlap across cameras makes cross-view matching difficult. However, when combined with temporal attention, view attention provides complementary cross-view information and further improves the shared representation.
Finally, adding Pl\"ucker ray embeddings injects explicit camera-aware geometric priors into the backbone, leading to the best performance across detection and occupancy metrics. These results demonstrate the effectiveness of adapting the VGGT backbone to streaming multi-view driving data.

\noindent \textbf{Number of Frames.}
Table~\ref{tab:frame_ablation} examines the effect of temporal context under two head configurations: current-only heads that decode current-frame features, and temporal heads with task-specific temporal modeling. All models are trained for 50 epochs. Increasing the number of input frames improves detection, occupancy, and depth, with most gains obtained by four frames and performance largely saturating thereafter. Temporal heads consistently outperform current-only heads, indicating that backbone-level multi-frame aggregation complements temporal modeling in downstream heads. Considering the trade-off between efficiency and performance, we use 4 input frames as the default setting in our experiments.

\begin{table}[t]
\centering
\footnotesize
\setlength{\tabcolsep}{3.5pt}
\renewcommand{\arraystretch}{1.05}
\begin{tabular}{l|ccc|cc|cc}
\hline
\multirow{2}{*}{Method} & \multicolumn{3}{c|}{Backbone Configuration} & \multicolumn{2}{c|}{Det.} & \multicolumn{2}{c}{Occ.} \\
& VGGT Blocks &  VGGT Init. &  Drive Adapt. & mAP$\uparrow$ & NDS$\uparrow$ & mIoU$\uparrow$ & RayIoU$\uparrow$ \\
\hline
DINOv2-L~\cite{DINOv2} & 0 & & & 51.9 & 59.7 & 34.2 & 38.8 \\
DINOv2-T (ours)  & 12 & & & 52.1 & 60.3 & 37.1 & 41.8 \\
VGGT-12~\cite{VGGT} & 12 & $\checkmark$ & & 54.6 & 62.0 & 38.8 & 42.6 \\
VGGT-24~\cite{VGGT} & 24 & $\checkmark$ & & \underline{55.5} & \underline{62.6} & \underline{39.6} & \underline{43.4} \\
\textbf{GeoUP (Ours)} & \textbf{12} & $\checkmark$ & $\checkmark$ & \textbf{56.4} & \textbf{63.0} & \textbf{40.3} & \textbf{44.5} \\
\hline
\end{tabular}
\caption{\textbf{Effect of backbone configurations} on the nuScenes validation set~\cite{nuScenes}. VGGT Blocks reports the number of VGGT-style Transformers blocks. VGGT Init. denotes initialization from geometry-pretrained VGGT weights, while Drive Adapt. identifies the proposed driving-oriented adaptations.}
\label{tab:baseline_comparison}
\end{table}

\begin{table}[t]
\centering
\scalebox{0.9}{
\begin{tabular}{cc|cc|cc}
\hline
\multirow{2}{*}{Occ. Sup.}
& \multirow{2}{*}{Det. Sup.}
& \multicolumn{2}{c|}{Det.}
& \multicolumn{2}{c}{Occ.} \\
&
& mAP$\uparrow$
& NDS$\uparrow$
& mIoU$\uparrow$
& RayIoU$\uparrow$ \\
\hline
&
$\checkmark$
& 55.9
& 62.8
& --
& -- \\

$\checkmark$
&
& --
& --
& 39.8
& 43.8 \\

$\checkmark$
& $\checkmark$
& \textbf{56.4}
& \textbf{63.0}
& \textbf{40.3}
& \textbf{44.5} \\
\hline
\end{tabular}
}
\caption{\textbf{Ablation study on multi-task learning} on the nuScenes validation set~\cite{nuScenes}.}
\label{tab:ablation_multitask}
\end{table}

\noindent \textbf{Effect of Backbone Configuration.}
Table~\ref{tab:baseline_comparison} compares different backbone configurations on the nuScenes validation set. DINOv2-L serves as the image-encoder-only reference,  while DINOv2-T augments it with 12 randomly initialized VGGT-style Transformers blocks. With the architecture and input configuration held fixed, initializing these blocks from pretrained VGGT weights (VGGT-12) improves mAP/NDS by 2.5/1.7 points and mIoU/RayIoU by 1.7/0.8 points. Increasing the number of VGGT-style blocks from 12 to 24 yields further improvements.
In comparison, GeoUP introduces the proposed driving-oriented adaptations on top of VGGT-12 and achieves the best results across all detection and occupancy metrics, even outperforming the deeper VGGT-24 backbone.

\begin{figure}[t]
\centering
\includegraphics[width=0.85\linewidth]{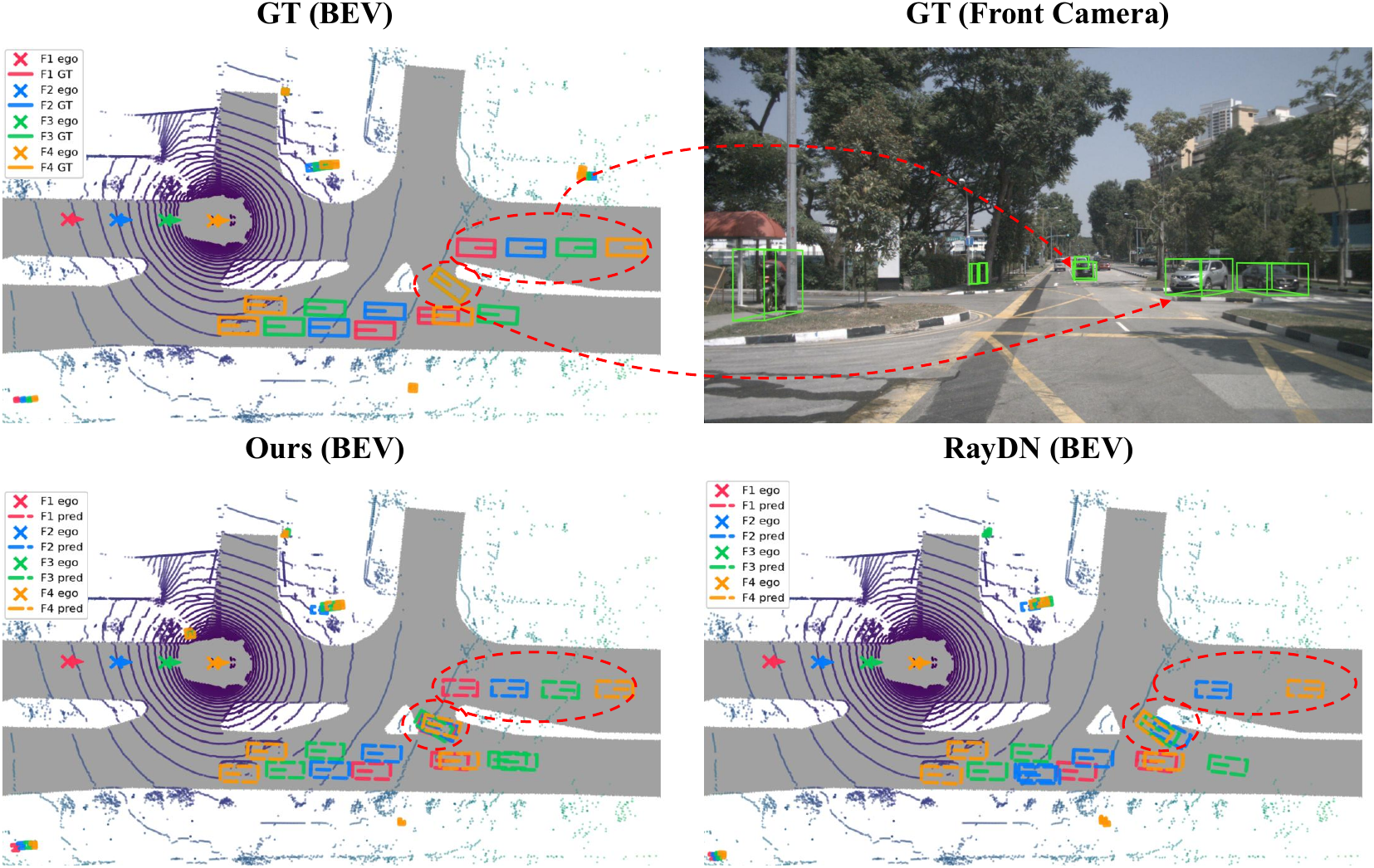}
\caption{
\textbf{Qualitative comparison of 3D detection results over consecutive frames.}
The red dashed regions indicate the corresponding vehicle areas in BEV, and the GT BEV region is linked to the associated vehicles in the front-camera image.
Compared with RayDN~\cite{RayDN}, GeoUP maintains more stable predictions for the static vehicle and continuously detects the moving vehicle across all four frames.
}
\label{vis_det}
\end{figure}

\begin{figure}[t]
\centering
\includegraphics[width=0.9\linewidth]{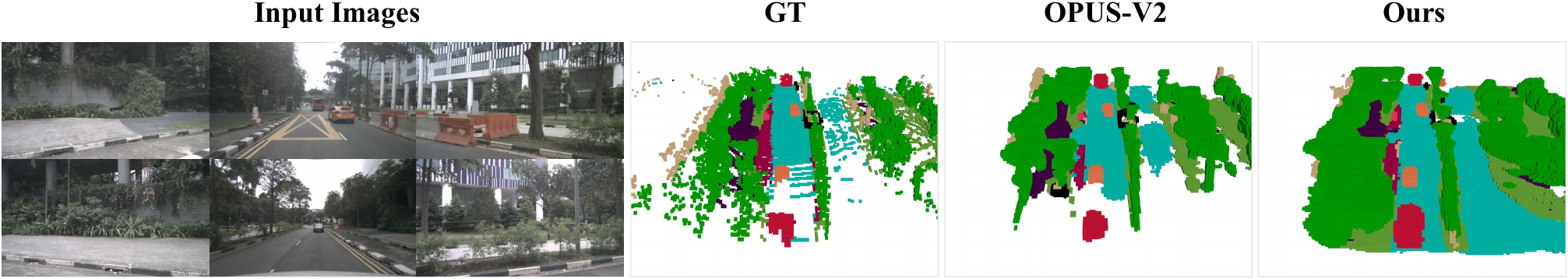}
\caption{
\textbf{Qualitative comparison of semantic occupancy prediction.} GeoUP better preserves the road layout and surrounding scene structures compared with OPUS-V2~\cite{OPUS}.
}
\label{vis_occ}
\end{figure}

\begin{figure}[t]
\centering
\includegraphics[width=0.85\linewidth]{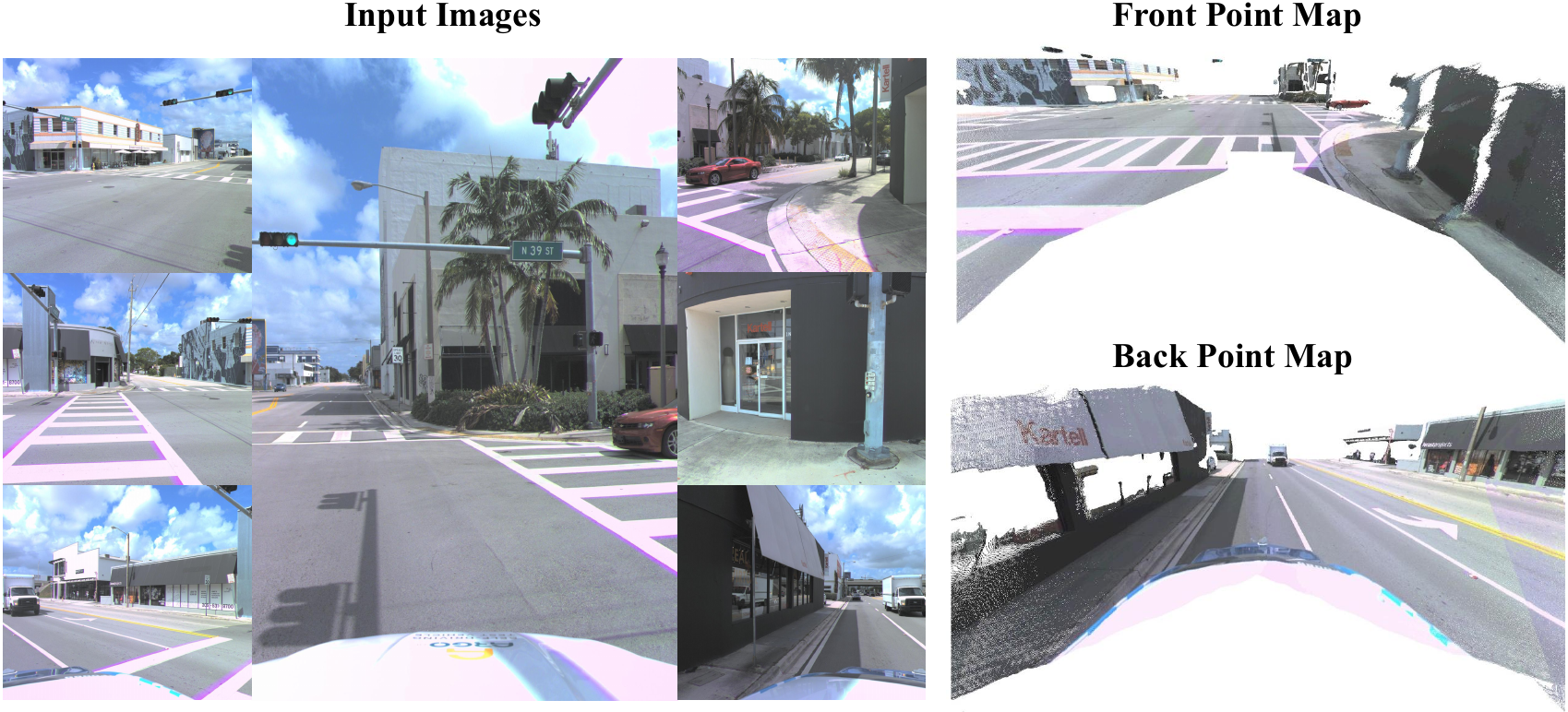}
\caption{
\textbf{Point maps reconstructed from the predicted depth.} GeoUP preserves the 3D reconstruction capability inherited from VGGT~\cite{VGGT}.
}
\label{vis_depth}
\end{figure}

\noindent \textbf{Effect of Multi-task Learning.}
Table~\ref{tab:ablation_multitask} studies the effect of joint optimization between occupancy prediction and 3D detection on the nuScenes validation set. Compared with training each task separately, the multi-task setting improves both detection and occupancy performance. These results suggest that multi-task learning encourages the backbone to learn a more complete and transferable 3D representation for autonomous driving perception.

\subsection{Qualitative Analysis}

We provide qualitative results across different perception tasks. For 3D detection, Figure~\ref{vis_det} compares GeoUP with RayDN~\cite{RayDN}. The red dashed regions mark the corresponding vehicle areas in BEV and link the GT BEV region to the front-camera image. For the highlighted static white vehicle, GeoUP maintains more stable box locations across frames. For the highlighted moving black vehicle, GeoUP detects it in all four frames, while RayDN misses it in two intermediate frames. These results show that the shared geometry-grounded representation can better aggregate historical observations and improve spatiotemporal consistency in 3D detection. 
For semantic occupancy prediction, Figure~\ref{vis_occ} compares GeoUP with OPUS-V2~\cite{OPUS}. GeoUP better preserves the continuous road layout and surrounding scene structures, producing more complete and coherent occupancy predictions. This suggests that the proposed geometry-grounded backbone provides stronger scene-level geometric and semantic understanding for occupancy prediction. For depth estimation, Figure~\ref{vis_depth} shows point maps reconstructed from the predicted depth. The clear road surfaces, vehicles, and surrounding structures indicate that GeoUP preserves the 3D reconstruction capability inherited from VGGT~\cite{VGGT}. Overall, these qualitative results demonstrate the effectiveness of GeoUP in producing accurate, coherent, and geometrically consistent perception outputs across detection, occupancy, and depth reconstruction. More comprehensive visualizations across all evaluated autonomous driving datasets and perception tasks are provided in the appendix.

\section{Conclusion and limitations} 
In this paper, we presented GeoUP, a geometry-grounded unified perception framework for autonomous driving. By adapting geometry foundation models to calibrated multi-camera driving scenes, GeoUP learns a shared 3D scene latent that supports metric depth estimation, 3D object detection, and semantic occupancy prediction. With temporal-view attention factorization, calibration-aware raymap injection, and joint multi-task and multi-dataset training, GeoUP transfers reconstruction-oriented representations to driving perception. Extensive experiments demonstrate state-of-the-art performance across perception tasks, validating geometry-grounded representation learning for unified autonomous driving perception.

Despite its strong performance, GeoUP has two main limitations. First, the visual geometry backbone is computationally heavy, leading to a large model size and limited inference speed. Efficient 3D reconstruction models, lightweight transformers, and model compression~\cite{ovggt,dinov3,fastvggt,infinitevggt,streamvggt} are promising directions for improvement. A detailed latency analysis is provided in the appendix. Second, the perception heads remain task-specific despite sharing the same geometry-grounded latent. Future work may explore a unified decoder for dense geometry, object instances, and semantic occupancy.

\section*{Acknowledgment}
This research is supported in part by  the National
Natural Science Foundation of China (No. 62461160308,
U23B2010, 62576024), the Beijing Natural Science Foun
dation (No. L231011), the Fundamental Research Funds
for the Central Universities (No. 501RCQD2025141003),
BeiHang GanWei Project (No. 502GWXM2024141001),
the National Science Foundation Support Projects (No.
62425303)
\bibliography{main}

\begin{thebibliography}{82}
\providecommand{\natexlab}[1]{#1}
\providecommand{\url}[1]{\texttt{#1}}
\expandafter\ifx\csname urlstyle\endcsname\relax
  \providecommand{\doi}[1]{doi: #1}\else
  \providecommand{\doi}{doi: \begingroup \urlstyle{rm}\Url}\fi

\bibitem[Bao et~al.(2021)Bao, Dong, Piao, and Wei]{BEiT}
Hangbo Bao, Li~Dong, Songhao Piao, and Furu Wei.
\newblock Beit: Bert pre-training of image transformers.
\newblock \emph{arXiv preprint arXiv:2106.08254}, 2021.

\bibitem[Cabon et~al.(2025)Cabon, Stoffl, Antsfeld, Csurka, Chidlovskii,
  Revaud, and Leroy]{MUSt3R}
Yohann Cabon, Lucas Stoffl, Leonid Antsfeld, Gabriela Csurka, Boris
  Chidlovskii, Jerome Revaud, and Vincent Leroy.
\newblock Must3r: Multi-view network for stereo 3d reconstruction.
\newblock In \emph{Proceedings of the IEEE/CVF Conference on Computer Vision
  and Pattern Recognition}, pages 1050--1060, 2025.

\bibitem[Caesar et~al.(2020)Caesar, Bankiti, Lang, Vora, Liong, Xu, Krishnan,
  Pan, Baldan, and Beijbom]{nuScenes}
Holger Caesar, Varun Bankiti, Alex~H Lang, Sourabh Vora, Venice~Erin Liong,
  Qiang Xu, Anush Krishnan, Yu~Pan, Giancarlo Baldan, and Oscar Beijbom.
\newblock nuscenes: A multimodal dataset for autonomous driving.
\newblock In \emph{Proceedings of the IEEE/CVF conference on computer vision
  and pattern recognition}, pages 11621--11631, 2020.

\bibitem[Cao et~al.(2025)Cao, Hallgarten, Li, Dauner, Gu, Wang, Miron, Aiello,
  Li, Gilitschenski, Ivanovic, Pavone, Geiger, and Chitta]{Cao2025CORL}
Wei Cao, Marcel Hallgarten, Tianyu Li, Daniel Dauner, Xunjiang Gu, Caojun Wang,
  Yakov Miron, Marco Aiello, Hongyang Li, Igor Gilitschenski, Boris Ivanovic,
  Marco Pavone, Andreas Geiger, and Kashyap Chitta.
\newblock Pseudo-simulation for autonomous driving.
\newblock In \emph{Conference on Robot Learning (CoRL)}, 2025.

\bibitem[Carion et~al.(2020)Carion, Massa, Synnaeve, Usunier, Kirillov, and
  Zagoruyko]{detr}
Nicolas Carion, Francisco Massa, Gabriel Synnaeve, Nicolas Usunier, Alexander
  Kirillov, and Sergey Zagoruyko.
\newblock End-to-end object detection with transformers.
\newblock In \emph{European conference on computer vision}, pages 213--229,
  2020.

\bibitem[Caron et~al.(2021)Caron, Touvron, Misra, J{\'e}gou, Mairal,
  Bojanowski, and Joulin]{DINO}
Mathilde Caron, Hugo Touvron, Ishan Misra, Herv{\'e} J{\'e}gou, Julien Mairal,
  Piotr Bojanowski, and Armand Joulin.
\newblock Emerging properties in self-supervised vision transformers.
\newblock In \emph{Proceedings of the IEEE/CVF international conference on
  computer vision}, pages 9650--9660, 2021.

\bibitem[Chen et~al.(2020)Chen, Kornblith, Norouzi, and Hinton]{SimCLR}
Ting Chen, Simon Kornblith, Mohammad Norouzi, and Geoffrey Hinton.
\newblock A simple framework for contrastive learning of visual
  representations.
\newblock In \emph{International conference on machine learning}, pages
  1597--1607, 2020.

\bibitem[D{\"a}hling et~al.(2026)D{\"a}hling, Krebs, and Z{\"o}llner]{DenseBEV}
Marius D{\"a}hling, Sebastian Krebs, and J~Marius Z{\"o}llner.
\newblock Densebev: Transforming bev grid cells into 3d objects.
\newblock In \emph{Proceedings of the IEEE/CVF Winter Conference on
  Applications of Computer Vision}, pages 2370--2379, 2026.

\bibitem[Dauner et~al.(2024)Dauner, Hallgarten, Li, Weng, Huang, Yang, Li,
  Gilitschenski, Ivanovic, Pavone, et~al.]{navsim}
Daniel Dauner, Marcel Hallgarten, Tianyu Li, Xinshuo Weng, Zhiyu Huang, Zetong
  Yang, Hongyang Li, Igor Gilitschenski, Boris Ivanovic, Marco Pavone, et~al.
\newblock Navsim: Data-driven non-reactive autonomous vehicle simulation and
  benchmarking.
\newblock \emph{Advances in Neural Information Processing Systems},
  37:\penalty0 28706--28719, 2024.

\bibitem[Deng et~al.(2009)Deng, Dong, Socher, Li, Li, and Fei-Fei]{imagenet}
Jia Deng, Wei Dong, Richard Socher, Li-Jia Li, Kai Li, and Li~Fei-Fei.
\newblock Imagenet: A large-scale hierarchical image database.
\newblock In \emph{2009 IEEE conference on computer vision and pattern
  recognition}, pages 248--255. Ieee, 2009.

\bibitem[Dosovitskiy et~al.(2020)Dosovitskiy, Beyer, Kolesnikov, Weissenborn,
  Zhai, Unterthiner, Dehghani, Minderer, Heigold, Gelly, et~al.]{ViT}
Alexey Dosovitskiy, Lucas Beyer, Alexander Kolesnikov, Dirk Weissenborn,
  Xiaohua Zhai, Thomas Unterthiner, Mostafa Dehghani, Matthias Minderer, Georg
  Heigold, Sylvain Gelly, et~al.
\newblock An image is worth 16x16 words: Transformers for image recognition at
  scale.
\newblock \emph{arXiv preprint arXiv:2010.11929}, 2020.

\bibitem[Duan et~al.(2019)Duan, Bai, Xie, Qi, Huang, and Tian]{CenterNet}
Kaiwen Duan, Song Bai, Lingxi Xie, Honggang Qi, Qingming Huang, and Qi~Tian.
\newblock Centernet: Keypoint triplets for object detection.
\newblock In \emph{Proceedings of the IEEE/CVF international conference on
  computer vision}, pages 6569--6578, 2019.

\bibitem[Fang et~al.(2023)Fang, Wang, Xie, Sun, Wu, Wang, Huang, Wang, and
  Cao]{EVA}
Yuxin Fang, Wen Wang, Binhui Xie, Quan Sun, Ledell Wu, Xinggang Wang, Tiejun
  Huang, Xinlong Wang, and Yue Cao.
\newblock Eva: Exploring the limits of masked visual representation learning at
  scale.
\newblock In \emph{Proceedings of the IEEE/CVF conference on computer vision
  and pattern recognition}, pages 19358--19369, 2023.

\bibitem[Fang et~al.(2024)Fang, Sun, Wang, Huang, Wang, and Cao]{EVA02}
Yuxin Fang, Quan Sun, Xinggang Wang, Tiejun Huang, Xinlong Wang, and Yue Cao.
\newblock Eva-02: A visual representation for neon genesis.
\newblock \emph{Image and Vision Computing}, 149:\penalty0 105171, 2024.

\bibitem[Geiger et~al.(2013)Geiger, Lenz, Stiller, and Urtasun]{KITTI}
Andreas Geiger, Philip Lenz, Christoph Stiller, and Raquel Urtasun.
\newblock Vision meets robotics: The kitti dataset.
\newblock \emph{The international journal of robotics research}, 32\penalty0
  (11):\penalty0 1231--1237, 2013.

\bibitem[Godard et~al.(2019)Godard, Mac~Aodha, Firman, and Brostow]{Monodepth2}
Cl{\'e}ment Godard, Oisin Mac~Aodha, Michael Firman, and Gabriel~J Brostow.
\newblock Digging into self-supervised monocular depth estimation.
\newblock In \emph{Proceedings of the IEEE/CVF International Conference on
  Computer Vision}, pages 3828--3838, 2019.

\bibitem[Guizilini et~al.(2020)Guizilini, Ambrus, Pillai, Raventos, and
  Gaidon]{DDAD}
Vitor Guizilini, Rares Ambrus, Sudeep Pillai, Allan Raventos, and Adrien
  Gaidon.
\newblock 3d packing for self-supervised monocular depth estimation.
\newblock In \emph{Proceedings of the IEEE/CVF conference on computer vision
  and pattern recognition}, pages 2485--2494, 2020.

\bibitem[He et~al.(2016)He, Zhang, Ren, and Sun]{ResNet}
Kaiming He, Xiangyu Zhang, Shaoqing Ren, and Jian Sun.
\newblock Deep residual learning for image recognition.
\newblock In \emph{Proceedings of the IEEE/CVF conference on computer vision
  and pattern recognition}, pages 770--778, 2016.

\bibitem[He et~al.(2020)He, Fan, Wu, Xie, and Girshick]{MoCo}
Kaiming He, Haoqi Fan, Yuxin Wu, Saining Xie, and Ross Girshick.
\newblock Momentum contrast for unsupervised visual representation learning.
\newblock In \emph{Proceedings of the IEEE/CVF conference on computer vision
  and pattern recognition}, pages 9729--9738, 2020.

\bibitem[He et~al.(2022)He, Chen, Xie, Li, Doll{\'a}r, and Girshick]{MAE}
Kaiming He, Xinlei Chen, Saining Xie, Yanghao Li, Piotr Doll{\'a}r, and Ross
  Girshick.
\newblock Masked autoencoders are scalable vision learners.
\newblock In \emph{Proceedings of the IEEE/CVF conference on computer vision
  and pattern recognition}, pages 16000--16009, 2022.

\bibitem[Huang et~al.(2021)Huang, Huang, Zhu, Ye, and Du]{BEVDet}
Junjie Huang, Guan Huang, Zheng Zhu, Yun Ye, and Dalong Du.
\newblock Bevdet: High-performance multi-camera 3d object detection in
  bird-eye-view.
\newblock \emph{arXiv preprint arXiv:2112.11790}, 2021.

\bibitem[Huang et~al.(2023)Huang, Zheng, Zhang, Zhou, and Lu]{TPVFormer}
Yuanhui Huang, Wenzhao Zheng, Yunpeng Zhang, Jie Zhou, and Jiwen Lu.
\newblock Tri-perspective view for vision-based 3d semantic occupancy
  prediction.
\newblock In \emph{Proceedings of the IEEE/CVF Conference on Computer Vision
  and Pattern Recognition}, pages 9223--9232, 2023.

\bibitem[Ji et~al.(2025)Ji, Ni, Huang, Shi, Luo, Zhan, and Chen]{RoPETR}
Hang Ji, Tao Ni, Xufeng Huang, Zhan Shi, Tao Luo, Xin Zhan, and Junbo Chen.
\newblock Ropetr: Improving temporal camera-only 3d detection by integrating
  enhanced rotary position embedding.
\newblock \emph{arXiv preprint arXiv:2504.12643}, 2025.

\bibitem[Jia et~al.(2025)Jia, Liu, You, Xia, Hong, and Yan]{drivevggt}
Xiaosong Jia, Yanhao Liu, Junqi You, Renqiu Xia, Yu~Hong, and Junchi Yan.
\newblock Drivevggt: Visual geometry transformer for autonomous driving.
\newblock \emph{arXiv preprint arXiv:2511.22264}, 2025.

\bibitem[Jiang et~al.(2024)Jiang, Li, Liu, Wang, Jia, Wang, Han, and
  Zhang]{far3d}
Xiaohui Jiang, Shuailin Li, Yingfei Liu, Shihao Wang, Fan Jia, Tiancai Wang,
  Lijin Han, and Xiangyu Zhang.
\newblock Far3d: Expanding the horizon for surround-view 3d object detection.
\newblock In \emph{Proceedings of the AAAI conference on artificial
  intelligence}, volume~38, pages 2561--2569, 2024.

\bibitem[Jordan et~al.(2024)Jordan, Jin, Boza, Jiacheng, Cesista, Newhouse, and
  Bernstein]{muon}
Keller Jordan, Yuchen Jin, Vlado Boza, You Jiacheng, Franz Cesista, Laker
  Newhouse, and Jeremy Bernstein.
\newblock Muon: An optimizer for hidden layers in neural networks, 2024.
\newblock \emph{URL https://kellerjordan. github. io/posts/muon}, 6\penalty0
  (3):\penalty0 4, 2024.

\bibitem[Keetha et~al.(2025)Keetha, M{\"u}ller, Sch{\"o}nberger, Porzi, Zhang,
  Fischer, Knapitsch, Zauss, Weber, Antunes, et~al.]{MapAnything}
Nikhil Keetha, Norman M{\"u}ller, Johannes Sch{\"o}nberger, Lorenzo Porzi,
  Yuchen Zhang, Tobias Fischer, Arno Knapitsch, Duncan Zauss, Ethan Weber,
  Nelson Antunes, et~al.
\newblock Mapanything: Universal feed-forward metric 3d reconstruction.
\newblock \emph{arXiv preprint arXiv:2509.13414}, 2025.

\bibitem[Lee et~al.(2019)Lee, Hwang, Lee, Bae, and Park]{vovnet-99}
Youngwan Lee, Joong-won Hwang, Sangrok Lee, Yuseok Bae, and Jongyoul Park.
\newblock An energy and gpu-computation efficient backbone network for
  real-time object detection.
\newblock In \emph{Proceedings of the IEEE/CVF conference on computer vision
  and pattern recognition workshops}, pages 0--0, 2019.

\bibitem[Leroy et~al.(2024)Leroy, Cabon, and Revaud]{MASt3R}
Vincent Leroy, Yohann Cabon, and J{\'e}r{\^o}me Revaud.
\newblock Grounding image matching in 3d with mast3r.
\newblock In \emph{European Conference on Computer Vision}, pages 71--91, 2024.

\bibitem[Li et~al.(2024{\natexlab{a}})Li, Huang, Wang, Rong, Wang, and
  Liu]{topo2d}
Han Li, Zehao Huang, Zitian Wang, Wenge Rong, Naiyan Wang, and Si~Liu.
\newblock Enhancing 3d lane detection and topology reasoning with 2d lane
  priors.
\newblock \emph{arXiv preprint arXiv:2406.03105}, 2024{\natexlab{a}}.

\bibitem[Li et~al.(2020)Li, Wang, Wu, Chen, Hu, Li, Tang, and Yang]{GFL}
Xiang Li, Wenhai Wang, Lijun Wu, Shuo Chen, Xiaolin Hu, Jun Li, Jinhui Tang,
  and Jian Yang.
\newblock Generalized focal loss: Learning qualified and distributed bounding
  boxes for dense object detection.
\newblock \emph{Advances in neural information processing systems},
  33:\penalty0 21002--21012, 2020.

\bibitem[Li et~al.(2023{\natexlab{a}})Li, Ge, Yu, Yang, Wang, Shi, Sun, and
  Li]{bevdepth}
Yinhao Li, Zheng Ge, Guanyi Yu, Jinrong Yang, Zengran Wang, Yukang Shi,
  Jianjian Sun, and Zeming Li.
\newblock Bevdepth: Acquisition of reliable depth for multi-view 3d object
  detection.
\newblock In \emph{Proceedings of the AAAI conference on artificial
  intelligence}, volume~37, pages 1477--1485, 2023{\natexlab{a}}.

\bibitem[Li et~al.(2023{\natexlab{b}})Li, Yu, Austin, Fang, Lan, Kautz, and
  Alvarez]{fb-occ}
Zhiqi Li, Zhiding Yu, David Austin, Mingsheng Fang, Shiyi Lan, Jan Kautz, and
  Jose~M Alvarez.
\newblock Fb-occ: 3d occupancy prediction based on forward-backward view
  transformation.
\newblock \emph{arXiv preprint arXiv:2307.01492}, 2023{\natexlab{b}}.

\bibitem[Li et~al.(2024{\natexlab{b}})Li, Wang, Li, Xie, Sima, Lu, Yu, and
  Dai]{BEVFormer}
Zhiqi Li, Wenhai Wang, Hongyang Li, Enze Xie, Chonghao Sima, Tong Lu, Qiao Yu,
  and Jifeng Dai.
\newblock Bevformer: learning bird’s-eye-view representation from
  lidar-camera via spatiotemporal transformers.
\newblock \emph{IEEE Transactions on Pattern Analysis and Machine
  Intelligence}, 47\penalty0 (3):\penalty0 2020--2036, 2024{\natexlab{b}}.

\bibitem[Lin et~al.(2017)Lin, Goyal, Girshick, He, and Doll{\'a}r]{FocalLoss}
Tsung-Yi Lin, Priya Goyal, Ross Girshick, Kaiming He, and Piotr Doll{\'a}r.
\newblock Focal loss for dense object detection.
\newblock In \emph{Proceedings of the IEEE international conference on computer
  vision}, pages 2980--2988, 2017.

\bibitem[Liu et~al.(2024{\natexlab{a}})Liu, Huang, Zhang, Yao, Zhang, Wan, Ye,
  and Zhou]{RayDN}
Feng Liu, Tengteng Huang, Qianjing Zhang, Haotian Yao, Chi Zhang, Fang Wan,
  Qixiang Ye, and Yanzhao Zhou.
\newblock Ray denoising: Depth-aware hard negative sampling for multi-view 3d
  object detection.
\newblock In \emph{European Conference on Computer Vision}, pages 200--217.
  Springer, 2024{\natexlab{a}}.

\bibitem[Liu et~al.(2024{\natexlab{b}})Liu, Chen, Wang, Yang, Li, Zeng, Chen,
  Li, and Wang]{SparseOcc}
Haisong Liu, Yang Chen, Haiguang Wang, Zetong Yang, Tianyu Li, Jia Zeng,
  Li~Chen, Hongyang Li, and Limin Wang.
\newblock Fully sparse 3d occupancy prediction.
\newblock In \emph{European Conference on Computer Vision}, pages 54--71.
  Springer, 2024{\natexlab{b}}.

\bibitem[Liu et~al.(2022{\natexlab{a}})Liu, Wang, Zhang, and Sun]{PETR}
Yingfei Liu, Tiancai Wang, Xiangyu Zhang, and Jian Sun.
\newblock Petr: Position embedding transformation for multi-view 3d object
  detection.
\newblock In \emph{European Conference on Computer Vision}, pages 531--548,
  2022{\natexlab{a}}.

\bibitem[Liu et~al.(2023{\natexlab{a}})Liu, Yan, Jia, Li, Gao, Wang, and
  Zhang]{PETRv2}
Yingfei Liu, Junjie Yan, Fan Jia, Shuailin Li, Aqi Gao, Tiancai Wang, and
  Xiangyu Zhang.
\newblock Petrv2: A unified framework for 3d perception from multi-camera
  images.
\newblock In \emph{Proceedings of the IEEE/CVF International Conference on
  Computer Vision}, pages 3262--3272, 2023{\natexlab{a}}.

\bibitem[Liu et~al.(2021)Liu, Lin, Cao, Hu, Wei, Zhang, Lin, and
  Guo]{SwinTransformer}
Ze~Liu, Yutong Lin, Yue Cao, Han Hu, Yixuan Wei, Zheng Zhang, Stephen Lin, and
  Baining Guo.
\newblock Swin transformer: Hierarchical vision transformer using shifted
  windows.
\newblock In \emph{Proceedings of the IEEE/CVF international conference on
  computer vision}, pages 10012--10022, 2021.

\bibitem[Liu et~al.(2023{\natexlab{b}})Liu, Tang, Amini, Yang, Mao, Rus, and
  Han]{BEVFusion}
Zhijian Liu, Haotian Tang, Alexander Amini, Xinyu Yang, Huizi Mao, Daniela~L
  Rus, and Song Han.
\newblock Bevfusion: Multi-task multi-sensor fusion with unified bird's-eye
  view representation.
\newblock In \emph{2023 IEEE international conference on robotics and
  automation (ICRA)}, pages 2774--2781. IEEE, 2023{\natexlab{b}}.

\bibitem[Liu et~al.(2022{\natexlab{b}})Liu, Mao, Wu, Feichtenhofer, Darrell,
  and Xie]{ConvNeXt}
Zhuang Liu, Hanzi Mao, Chao-Yuan Wu, Christoph Feichtenhofer, Trevor Darrell,
  and Saining Xie.
\newblock A convnet for the 2020s.
\newblock In \emph{Proceedings of the IEEE/CVF conference on computer vision
  and pattern recognition}, pages 11976--11986, 2022{\natexlab{b}}.

\bibitem[Loshchilov and Hutter(2016)]{cos}
Ilya Loshchilov and Frank Hutter.
\newblock Sgdr: Stochastic gradient descent with warm restarts.
\newblock \emph{arXiv preprint arXiv:1608.03983}, 2016.

\bibitem[Loshchilov and Hutter(2017)]{AdamW}
Ilya Loshchilov and Frank Hutter.
\newblock Decoupled weight decay regularization.
\newblock \emph{arXiv preprint arXiv:1711.05101}, 2017.

\bibitem[Lu et~al.(2026)Lu, Chen, Hsu, Jhong, Cheng, and Chen]{ovggt}
Si-Yu Lu, Po-Ting Chen, Hui-Che Hsu, Sin-Ye Jhong, Wen-Huang Cheng, and
  Yung-Yao Chen.
\newblock Ovggt: O(1) constant-cost streaming visual geometry transformer.
\newblock \emph{arXiv preprint arXiv:2603.05959}, 2026.

\bibitem[Oquab et~al.(2023)Oquab, Darcet, Moutakanni, Vo, Szafraniec, Khalidov,
  Fernandez, Haziza, Massa, El-Nouby, et~al.]{DINOv2}
Maxime Oquab, Timoth{\'e}e Darcet, Th{\'e}o Moutakanni, Huy Vo, Marc
  Szafraniec, Vasil Khalidov, Pierre Fernandez, Daniel Haziza, Francisco Massa,
  Alaaeldin El-Nouby, et~al.
\newblock Dinov2: Learning robust visual features without supervision.
\newblock \emph{arXiv preprint arXiv:2304.07193}, 2023.

\bibitem[Park et~al.(2021)Park, Ambrus, Guizilini, Li, and Gaidon]{DD3D}
Dennis Park, Rares Ambrus, Vitor Guizilini, Jie Li, and Adrien Gaidon.
\newblock Is pseudo-lidar needed for monocular 3d object detection?
\newblock In \emph{Proceedings of the IEEE/CVF international conference on
  computer vision}, pages 3142--3152, 2021.

\bibitem[Philion and Fidler(2020)]{LSS}
Jonah Philion and Sanja Fidler.
\newblock Lift, splat, shoot: Encoding images from arbitrary camera rigs by
  implicitly unprojecting to 3d.
\newblock In \emph{European conference on computer vision}, 2020.

\bibitem[Radford et~al.(2021)Radford, Kim, Hallacy, Ramesh, Goh, Agarwal,
  Sastry, Askell, Mishkin, Clark, et~al.]{CLIP}
Alec Radford, Jong~Wook Kim, Chris Hallacy, Aditya Ramesh, Gabriel Goh,
  Sandhini Agarwal, Girish Sastry, Amanda Askell, Pamela Mishkin, Jack Clark,
  et~al.
\newblock Learning transferable visual models from natural language
  supervision.
\newblock In \emph{International conference on machine learning}, pages
  8748--8763. PMLR, 2021.

\bibitem[Ranftl et~al.(2021)Ranftl, Bochkovskiy, and Koltun]{DPT}
Ren{\'e} Ranftl, Alexey Bochkovskiy, and Vladlen Koltun.
\newblock Vision transformers for dense prediction.
\newblock In \emph{Proceedings of the IEEE/CVF international conference on
  computer vision}, pages 12179--12188, 2021.

\bibitem[Sarlin et~al.(2020)Sarlin, DeTone, Malisiewicz, and
  Rabinovich]{SuperGlue}
Paul-Edouard Sarlin, Daniel DeTone, Tomasz Malisiewicz, and Andrew Rabinovich.
\newblock Superglue: Learning feature matching with graph neural networks.
\newblock In \emph{Proceedings of the IEEE/CVF conference on computer vision
  and pattern recognition}, pages 4938--4947, 2020.

\bibitem[Schonberger and Frahm(2016)]{COLMAP}
Johannes~L Schonberger and Jan-Michael Frahm.
\newblock Structure-from-motion revisited.
\newblock In \emph{Proceedings of the IEEE conference on computer vision and
  pattern recognition}, pages 4104--4113, 2016.

\bibitem[Shen et~al.(2025)Shen, Zhang, Qu, Zheng, Ji, Zhang, and Cao]{fastvggt}
You Shen, Zhipeng Zhang, Yansong Qu, Xiawu Zheng, Jiayi Ji, Shengchuan Zhang,
  and Liujuan Cao.
\newblock Fastvggt: Training-free acceleration of visual geometry transformer.
\newblock \emph{arXiv preprint arXiv:2509.02560}, 2025.

\bibitem[Shu et~al.(2023)Shu, Deng, Yu, and Liu]{3DPPE}
Changyong Shu, Jiajun Deng, Fisher Yu, and Yifan Liu.
\newblock 3dppe: 3d point positional encoding for transformer-based
  multi-camera 3d object detection.
\newblock In \emph{Proceedings of the IEEE/CVF International Conference on
  Computer Vision}, pages 3580--3589, 2023.

\bibitem[Sim{\'e}oni et~al.(2025)Sim{\'e}oni, Vo, Seitzer, Baldassarre, Oquab,
  Jose, Khalidov, Szafraniec, Yi, Ramamonjisoa, et~al.]{dinov3}
Oriane Sim{\'e}oni, Huy~V Vo, Maximilian Seitzer, Federico Baldassarre, Maxime
  Oquab, Cijo Jose, Vasil Khalidov, Marc Szafraniec, Seungeun Yi, Micha{\"e}l
  Ramamonjisoa, et~al.
\newblock Dinov3.
\newblock \emph{arXiv preprint arXiv:2508.10104}, 2025.

\bibitem[Sitzmann et~al.(2021)Sitzmann, Rezchikov, Freeman, Tenenbaum, and
  Durand]{light}
Vincent Sitzmann, Semon Rezchikov, Bill Freeman, Josh Tenenbaum, and Fredo
  Durand.
\newblock Light field networks: Neural scene representations with
  single-evaluation rendering.
\newblock \emph{Advances in Neural Information Processing Systems},
  34:\penalty0 19313--19325, 2021.

\bibitem[Sun et~al.(2021)Sun, Shen, Wang, Bao, and Zhou]{LoFTR}
Jiaming Sun, Zehong Shen, Yuang Wang, Hujun Bao, and Xiaowei Zhou.
\newblock Loftr: Detector-free local feature matching with transformers.
\newblock In \emph{Proceedings of the IEEE/CVF conference on computer vision
  and pattern recognition}, pages 8922--8931, 2021.

\bibitem[Sun et~al.(2020)Sun, Kretzschmar, Dotiwalla, Chouard, Patnaik, Tsui,
  Guo, Zhou, Chai, Caine, et~al.]{Waymo}
Pei Sun, Henrik Kretzschmar, Xerxes Dotiwalla, Aurelien Chouard, Vijaysai
  Patnaik, Paul Tsui, James Guo, Yin Zhou, Yuning Chai, Benjamin Caine, et~al.
\newblock Scalability in perception for autonomous driving: Waymo open dataset.
\newblock In \emph{Proceedings of the IEEE/CVF conference on computer vision
  and pattern recognition}, pages 2446--2454, 2020.

\bibitem[Tian et~al.(2023)Tian, Jiang, Yun, Mao, Yang, Wang, Wang, and
  Zhao]{occ3d}
Xiaoyu Tian, Tao Jiang, Longfei Yun, Yucheng Mao, Huitong Yang, Yue Wang, Yilun
  Wang, and Hang Zhao.
\newblock Occ3d: A large-scale 3d occupancy prediction benchmark for autonomous
  driving.
\newblock \emph{Advances in Neural Information Processing Systems},
  36:\penalty0 64318--64330, 2023.

\bibitem[Wang et~al.(2021{\natexlab{a}})Wang, Galliani, Vogel, Speciale, and
  Pollefeys]{PatchmatchNet}
Fangjinhua Wang, Silvano Galliani, Christoph Vogel, Pablo Speciale, and Marc
  Pollefeys.
\newblock Patchmatchnet: Learned multi-view patchmatch stereo.
\newblock In \emph{Proceedings of the IEEE/CVF conference on computer vision
  and pattern recognition}, pages 14194--14203, 2021{\natexlab{a}}.

\bibitem[Wang et~al.(2024{\natexlab{a}})Wang, Liu, Meng, Yan, Wang, Yang, Liu,
  Hou, and Cheng]{OPUS}
Jiabao Wang, Zhaojiang Liu, Qiang Meng, Liujiang Yan, Ke~Wang, Jie Yang, Wei
  Liu, Qibin Hou, and Ming-Ming Cheng.
\newblock Opus: occupancy prediction using a sparse set.
\newblock \emph{Advances in Neural Information Processing Systems},
  37:\penalty0 119861--119885, 2024{\natexlab{a}}.

\bibitem[Wang et~al.(2024{\natexlab{b}})Wang, Karaev, Rupprecht, and
  Novotny]{VGGSfM}
Jianyuan Wang, Nikita Karaev, Christian Rupprecht, and David Novotny.
\newblock Vggsfm: Visual geometry grounded deep structure from motion.
\newblock In \emph{Proceedings of the IEEE/CVF Conference on Computer Vision
  and Pattern Recognition}, pages 21686--21697, 2024{\natexlab{b}}.

\bibitem[Wang et~al.(2025{\natexlab{a}})Wang, Chen, Karaev, Vedaldi, Rupprecht,
  and Novotny]{VGGT}
Jianyuan Wang, Minghao Chen, Nikita Karaev, Andrea Vedaldi, Christian
  Rupprecht, and David Novotny.
\newblock Vggt: Visual geometry grounded transformer.
\newblock In \emph{Proceedings of the IEEE/CVF Conference on Computer Vision
  and Pattern Recognition}, pages 5294--5306, 2025{\natexlab{a}}.

\bibitem[Wang et~al.(2023{\natexlab{a}})Wang, Liu, Wang, Li, and
  Zhang]{StreamPETR}
Shihao Wang, Yingfei Liu, Tiancai Wang, Ying Li, and Xiangyu Zhang.
\newblock Exploring object-centric temporal modeling for efficient multi-view
  3d object detection.
\newblock In \emph{Proceedings of the IEEE/CVF International Conference on
  Computer Vision}, pages 3621--3631, 2023{\natexlab{a}}.

\bibitem[Wang et~al.(2024{\natexlab{c}})Wang, Leroy, Cabon, Chidlovskii, and
  Revaud]{DUSt3R}
Shuzhe Wang, Vincent Leroy, Yohann Cabon, Boris Chidlovskii, and Jerome Revaud.
\newblock Dust3r: Geometric 3d vision made easy.
\newblock In \emph{Proceedings of the IEEE/CVF Conference on Computer Vision
  and Pattern Recognition}, pages 20697--20709, 2024{\natexlab{c}}.

\bibitem[Wang et~al.(2021{\natexlab{b}})Wang, Zhu, Pang, and Lin]{FCOS3D}
Tai Wang, Xinge Zhu, Jiangmiao Pang, and Dahua Lin.
\newblock Fcos3d: Fully convolutional one-stage monocular 3d object detection.
\newblock In \emph{Proceedings of the IEEE/CVF international conference on
  computer vision}, pages 913--922, 2021{\natexlab{b}}.

\bibitem[Wang et~al.(2022{\natexlab{a}})Wang, Lian, Zhu, Zhu, and
  Zhang]{MV-Fcos3D++}
Tai Wang, Qing Lian, Chenming Zhu, Xinge Zhu, and Wenwei Zhang.
\newblock Mv-fcos3d++: Multi-view camera-only 4d object detection with
  pretrained monocular backbones.
\newblock \emph{arXiv preprint arXiv:2207.12716}, 2022{\natexlab{a}}.

\bibitem[Wang et~al.(2025{\natexlab{b}})Wang, Zhou, Zhu, Chang, Zhou, Li, Chen,
  Pang, Shen, and He]{pi3}
Yifan Wang, Jianjun Zhou, Haoyi Zhu, Wenzheng Chang, Yang Zhou, Zizun Li, Junyi
  Chen, Jiangmiao Pang, Chunhua Shen, and Tong He.
\newblock $\pi^3$: Permutation-equivariant visual geometry learning.
\newblock \emph{arXiv preprint arXiv:2507.13347}, 2025{\natexlab{b}}.

\bibitem[Wang et~al.(2022{\natexlab{b}})Wang, Guizilini, Zhang, Wang, Zhao, and
  Solomon]{DETR3D}
Yue Wang, Vitor~Campagnolo Guizilini, Tianyuan Zhang, Yilun Wang, Hang Zhao,
  and Justin Solomon.
\newblock Detr3d: 3d object detection from multi-view images via 3d-to-2d
  queries.
\newblock In \emph{Conference on robot learning}, pages 180--191,
  2022{\natexlab{b}}.

\bibitem[Wang et~al.(2023{\natexlab{b}})Wang, Huang, Fu, Wang, and Liu]{mv2d}
Zitian Wang, Zehao Huang, Jiahui Fu, Naiyan Wang, and Si~Liu.
\newblock Object as query: Lifting any 2d object detector to 3d detection.
\newblock In \emph{Proceedings of the IEEE/CVF International Conference on
  Computer Vision}, pages 3791--3800, 2023{\natexlab{b}}.

\bibitem[Wei et~al.(2023)Wei, Zhao, Zheng, Zhu, Zhou, and Lu]{SurroundOcc}
Yi~Wei, Linqing Zhao, Wenzhao Zheng, Zheng Zhu, Jie Zhou, and Jiwen Lu.
\newblock Surroundocc: Multi-camera 3d occupancy prediction for autonomous
  driving.
\newblock In \emph{Proceedings of the IEEE/CVF International Conference on
  Computer Vision}, pages 21729--21740, 2023.

\bibitem[Wilson et~al.(2023)Wilson, Qi, Agarwal, Lambert, Singh, Khandelwal,
  Pan, Kumar, Hartnett, Pontes, et~al.]{Argoverse2}
Benjamin Wilson, William Qi, Tanmay Agarwal, John Lambert, Jagjeet Singh,
  Siddhesh Khandelwal, Bowen Pan, Ratnesh Kumar, Andrew Hartnett,
  Jhony~Kaesemodel Pontes, et~al.
\newblock Argoverse 2: Next generation datasets for self-driving perception and
  forecasting.
\newblock \emph{arXiv preprint arXiv:2301.00493}, 2023.

\bibitem[Yang et~al.(2023)Yang, Chen, Tian, Tao, Zhu, Zhang, Huang, Li, Qiao,
  Lu, et~al.]{BEVFormerV2}
Chenyu Yang, Yuntao Chen, Hao Tian, Chenxin Tao, Xizhou Zhu, Zhaoxiang Zhang,
  Gao Huang, Hongyang Li, Yu~Qiao, Lewei Lu, et~al.
\newblock Bevformer v2: Adapting modern image backbones to bird's-eye-view
  recognition via perspective supervision.
\newblock In \emph{Proceedings of the IEEE/CVF conference on computer vision
  and pattern recognition}, pages 17830--17839, 2023.

\bibitem[Yang et~al.(2024)Yang, Kang, Huang, Xu, Feng, and Zhao]{yang2024depth}
Lihe Yang, Bingyi Kang, Zilong Huang, Xiaogang Xu, Jiashi Feng, and Hengshuang
  Zhao.
\newblock Depth anything: Unleashing the power of large-scale unlabeled data.
\newblock In \emph{2024 IEEE/CVF Conference on Computer Vision and Pattern
  Recognition (CVPR)}, pages 10371--10381. IEEE, 2024.

\bibitem[Yao et~al.(2026)Yao, Li, Lan, Wang, Sun, Alvarez, and Wu]{drivesuprim}
Wenhao Yao, Zhenxin Li, Shiyi Lan, Zi~Wang, Xinglong Sun, Jose~M Alvarez, and
  Zuxuan Wu.
\newblock Drivesuprim: Towards precise trajectory selection for end-to-end
  planning.
\newblock In \emph{Proceedings of the AAAI Conference on Artificial
  Intelligence}, volume~40, pages 11910--11918, 2026.

\bibitem[Yao et~al.(2018)Yao, Luo, Li, Fang, and Quan]{MVSNet}
Yao Yao, Zixin Luo, Shiwei Li, Tian Fang, and Long Quan.
\newblock Mvsnet: Depth inference for unstructured multi-view stereo.
\newblock In \emph{Proceedings of the European conference on computer vision
  (ECCV)}, pages 767--783, 2018.

\bibitem[Yuan et~al.(2026)Yuan, Yang, Yang, Zhang, Zhao, Zhang, and
  Zhang]{infinitevggt}
Shuai Yuan, Yantai Yang, Xiaotian Yang, Xupeng Zhang, Zhonghao Zhao, Lingming
  Zhang, and Zhipeng Zhang.
\newblock Infinitevggt: Visual geometry grounded transformer for endless
  streams.
\newblock \emph{arXiv preprint arXiv:2601.02281}, 2026.

\bibitem[Zhai et~al.(2023)Zhai, Mustafa, Kolesnikov, and Beyer]{SigLIP}
Xiaohua Zhai, Basil Mustafa, Alexander Kolesnikov, and Lucas Beyer.
\newblock Sigmoid loss for language image pre-training.
\newblock In \emph{Proceedings of the IEEE/CVF international conference on
  computer vision}, pages 11975--11986, 2023.

\bibitem[Zhang et~al.(2023)Zhang, Zhu, and Du]{OccFormer}
Yunpeng Zhang, Zheng Zhu, and Dalong Du.
\newblock Occformer: Dual-path transformer for vision-based 3d semantic
  occupancy prediction.
\newblock In \emph{Proceedings of the IEEE/CVF International Conference on
  Computer Vision}, pages 9433--9443, 2023.

\bibitem[Zheng and Vedaldi(2024)]{free3d}
Chuanxia Zheng and Andrea Vedaldi.
\newblock Free3d: Consistent novel view synthesis without 3d representation.
\newblock In \emph{Proceedings of the IEEE/CVF Conference on Computer Vision
  and Pattern Recognition}, pages 9720--9731, 2024.

\bibitem[Zhuo et~al.(2025)Zhuo, Zheng, Guo, Wu, Zhou, and Lu]{streamvggt}
Dong Zhuo, Wenzhao Zheng, Jiahe Guo, Yuqi Wu, Jie Zhou, and Jiwen Lu.
\newblock Streaming 4d visual geometry transformer.
\newblock \emph{arXiv preprint arXiv:2507.11539}, 2025.

\bibitem[Zuo et~al.(2025)Zuo, Xie, Zheng, Xu, Li, Jiang, Chen, Yang, and
  Lu]{dvgt}
Sicheng Zuo, Zixun Xie, Wenzhao Zheng, Shaoqing Xu, Fang Li, Shengyin Jiang,
  Long Chen, Zhi-Xin Yang, and Jiwen Lu.
\newblock Dvgt: Driving visual geometry transformer.
\newblock \emph{arXiv preprint arXiv:2512.16919}, 2025.

\end{thebibliography}

\clearpage
\appendix
\setcounter{page}{1}
\section*{Appendix}

\section{Additional Dataset and Implementation Details}
\noindent\textbf{Datasets and evaluation protocols.}
We provide additional details of the datasets and evaluation protocols used in the main paper.

\textbf{nuScenes and Occ3D-nuScenes.} nuScenes~\cite{nuScenes} contains 1,000 driving scenes captured by six cameras with 360$^\circ$ surround-view coverage and provides 3D annotations for 10 object categories. For 3D detection, we follow the official protocol and report mean Average Precision (mAP), nuScenes Detection Score (NDS), and five true-positive metrics: Average Translation Error (mATE), Average Scale Error (mASE), Average Orientation Error (mAOE), Average Velocity Error (mAVE), and Average Attribute Error (mAAE). For occupancy prediction, we evaluate on Occ3D-nuScenes~\cite{occ3d}, which provides annotations for 18 semantic occupancy classes. We report mean Intersection-over-Union (mIoU) and Ray-level mIoU (RayIoU)~\cite{SparseOcc}, including RayIoU under the 1\,m, 2\,m, and 4\,m thresholds.

\textbf{Waymo Open Dataset.} Waymo~\cite{Waymo} provides large-scale driving sequences captured by five cameras covering the front and side views. Following common camera-only 3D detection practice, we use 20\% of the training split and report the official Longitudinal Error Tolerant 3D Average Precision (LET-3D-AP), its heading-accuracy-weighted variant (LET-3D-APH), and its longitudinal-affinity-weighted variant (LET-3D-APL).

\textbf{Argoverse 2.} Argoverse 2~\cite{Argoverse2} provides 360$^\circ$ multi-camera driving data captured by seven ring cameras, together with additional stereo cameras, and contains 3D annotations for 26 object categories. We follow the official 3D detection protocol and report mAP, Composite Detection Score (CDS), Average Translation Error (mATE), Average Scale Error (mASE), and Average Orientation Error (mAOE).

\textbf{KITTI.} KITTI~\cite{KITTI} provides forward-view driving data captured by a front-facing stereo camera setup. We use it to evaluate monocular depth transferability to forward-view scenes and report Absolute Relative Error (Abs Rel) and threshold accuracy $\delta<1.25$.

\textbf{DDAD.} DDAD~\cite{DDAD} provides high-resolution driving data captured by six synchronized cameras with 360$^\circ$ surround-view coverage. We use it to evaluate dense metric depth prediction and cross-dataset geometric generalization, using the same Abs Rel and $\delta<1.25$ metrics as KITTI.

\noindent\textbf{Multi-dataset configuration.}
For multi-dataset training, GeoUP is trained on nuScenes~\cite{nuScenes}, Argoverse 2~\cite{Argoverse2}, Waymo~\cite{Waymo}, DDAD~\cite{DDAD}, and KITTI~\cite{KITTI}. 
These datasets provide heterogeneous annotations for different tasks. For depth estimation and camera pose prediction, all datasets are used for supervision, and the depth targets are normalized by a unified depth scale to maintain consistent metric supervision across datasets, where we set the depth scale to 90\,m. For 3D detection, we mainly use nuScenes~\cite{nuScenes}, Waymo~\cite{Waymo}, and Argoverse~2~\cite{Argoverse2}, where all 3D boxes are represented under a nuScenes-style~\cite{nuScenes} LiDAR coordinate convention. Since the perception scale differs across datasets, we use dataset-specific point cloud ranges, with maximum horizontal ranges of 51.2\,m, 74.88\,m, and 152.4\,m for nuScenes~\cite{nuScenes}, Waymo~\cite{Waymo}, and Argoverse~2~\cite{Argoverse2}, respectively. The detection box regression branch is shared across datasets, while the classification heads are separated according to dataset-specific category spaces. For occupancy prediction, supervision is only available on nuScenes~\cite{nuScenes}, so the occupancy branch follows the nuScenes occupancy configuration and is optimized only on nuScenes samples.

\noindent\textbf{Loss functions.}
GeoUP is optimized with task-specific objectives for depth estimation, 3D object detection, occupancy prediction, and camera pose prediction. 
For depth estimation, the DPT-style~\cite{DPT} head predicts a dense depth map and is trained with a depth objective that combines direct depth regression and gradient-based regularization.
The direct regression term supervises metric depth values with an L2 objective, and the gradient term penalizes image-space depth-gradient errors to improve local geometric consistency. 
The depth loss is written as:
\begin{equation}
\mathcal{L}_{\mathrm{dep}}
=
1.0\cdot\mathcal{L}_{\mathrm{reg}}
+
1.0\cdot\mathcal{L}_{\mathrm{grad}},
\end{equation}
where $\mathcal{L}_{\mathrm{reg}}$ and $\mathcal{L}_{\mathrm{grad}}$ denote the direct depth regression loss and the gradient regularization loss, respectively.

For 3D object detection, the RayDN-style~\cite{RayDN} detection head follows DETR-style~\cite{detr} Hungarian assignment and denoising training. 
The matching cost uses focal classification cost~\cite{FocalLoss} and L1 box regression cost. The detection objective contains focal classification loss~\cite{FocalLoss}, L1 box regression loss, and denoising loss:
\begin{equation}
\mathcal{L}_{\mathrm{det}}
=
2.0\cdot\mathcal{L}_{\mathrm{cls}}^{3\mathrm{D}}
+
0.25\cdot\mathcal{L}_{\mathrm{box}}^{3\mathrm{D}}
+
1.0\cdot\mathcal{L}_{\mathrm{dn}}.
\end{equation}
The focal loss~\cite{FocalLoss} uses $\gamma=2.0$ and $\alpha=0.25$.

We also keep an auxiliary 2D head during training to provide image-space localization supervision. 
This branch uses Quality Focal Loss~\cite{GFL} for 2D classification,
Gaussian focal loss~\cite{CenterNet} for centerness prediction, L1 loss for 2D box regression, GIoU loss for 2D box overlap, and L1 loss for projected center regression:
\begin{equation}
\mathcal{L}_{2\mathrm{D}}
=
2.0\cdot\mathcal{L}_{\mathrm{qfl}}
+
1.0\cdot\mathcal{L}_{\mathrm{center}}
+
5.0\cdot\mathcal{L}_{\mathrm{box}}^{2\mathrm{D}}
+
2.0\cdot\mathcal{L}_{\mathrm{giou}}^{2\mathrm{D}}
+
10.0\cdot\mathcal{L}_{\mathrm{center2d}} .
\end{equation}

For occupancy prediction, the OPUS-V2-style~\cite{OPUS} head predicts sparse occupied points and their semantic labels. 
The semantic branch is supervised by focal classification loss, and the point branch is supervised by Smooth-L1 regression loss with $\beta=0.2$. 
The occupancy objective is:
\begin{equation}
\mathcal{L}_{\mathrm{occ}}
=
2.0\,\mathcal{L}_{\mathrm{cls}}^{\mathrm{occ}}
+
0.5\,\mathcal{L}_{\mathrm{pts}}^{\mathrm{occ}}.
\end{equation}

The camera head follows the VGGT-style~\cite{VGGT} camera representation and is supervised with an L1 camera loss:
\begin{equation}
\mathcal{L}_{\mathrm{cam}}
=
5.0\cdot\mathcal{L}_{1}^{\mathrm{cam}}.
\end{equation}

For multi-task and multi-dataset training, we apply supervision masks according to annotation availability. 
The final training objective is:
\begin{equation}
\mathcal{L}
=
m_{\mathrm{dep}}\lambda_{\mathrm{dep}}\mathcal{L}_{\mathrm{dep}}
+
m_{\mathrm{det}}\mathcal{L}_{\mathrm{det}}
+
m_{2\mathrm{D}}\mathcal{L}_{2\mathrm{D}}
+
m_{\mathrm{occ}}\mathcal{L}_{\mathrm{occ}}
+
m_{\mathrm{cam}}\lambda_{\mathrm{cam}}\mathcal{L}_{\mathrm{cam}},
\end{equation}
where $m_{\mathrm{dep}}$, $m_{\mathrm{det}}$, $m_{2\mathrm{D}}$, $m_{\mathrm{occ}}$, and $m_{\mathrm{cam}}$ indicate whether the corresponding supervision is available for the sampled data. 
Following the main training setting, the geometry-oriented depth and camera objectives are down-weighted in the unified stage with $\lambda_{\mathrm{dep}}=\lambda_{\mathrm{cam}}=0.1$.
\section{Efficiency Analysis}

\begin{table}[t]
\centering
\small
\begin{tabular}{l|c|c|c}
\hline
Method & Backbone & Image Size & FPS $\uparrow$ \\
\hline
RayDN~\cite{RayDN} & EVA02 & $672\times224$ & 9.94 \\
OPUS-V2~\cite{OPUS} & ViT-L & $672\times224$ & 5.90 \\
\hline
GeoUP (Ours, 1f) & ViT-L & $672\times224$ & 2.18 \\
GeoUP (Ours, 4f) & ViT-L & $672\times224$ & 0.81 \\
\hline
\end{tabular}

\caption{\textbf{FPS comparison} with task-specific baselines at $672\times224$ resolution.}
\label{tab:appendix_efficiency_comparison}
\end{table}

We further analyze the inference efficiency of GeoUP. 
All runtime results are measured on a single NVIDIA H20 GPU with batch size 1 after 20 warm-up iterations and 100 inference iterations. 
For this profiling experiment, we use an input resolution of $672\times224$. 
The reported FPS is computed from the average end-to-end runtime. 
Since GeoUP jointly predicts depth, 3D detection, occupancy, and camera pose, its runtime includes the shared geometry backbone and all task readout heads. 
For reference, we also compare with task-specific RayDN~\cite{RayDN} and OPUS-V2~\cite{OPUS} models. 
As shown in Table~\ref{tab:appendix_efficiency_comparison}, GeoUP is slower than task-specific perception models because it uses a VGGT-style~\cite{VGGT} geometry backbone and produces multiple perception outputs in a single forward pass. 
When the temporal window increases from 1 frame to 4 frames, the FPS decreases from 2.18 to 0.81. 
This indicates that the main efficiency bottleneck comes from multi-frame geometry modeling in the shared backbone, rather than from the downstream task heads.

\begin{table}[t]
\centering
\small
\begin{tabular}{l|cc|cc}
\hline
\multirow{2}{*}{Module} 
& \multicolumn{2}{c|}{1 frame} 
& \multicolumn{2}{c}{4 frames} \\
& Latency (ms) & Ratio (\%) & Latency (ms) & Ratio (\%) \\
\hline
Geometry-grounded backbone & 292.7 & 66.5 & 1062.1 & 87.8 \\
Detection head & 28.5 & 6.5 & 28.7 & 2.4 \\
Depth head & 32.5 & 7.4 & 32.5 & 2.7 \\
Occupancy head & 81.5 & 18.5 & 81.6 & 6.7 \\
Camera head & 5.1 & 1.2 & 5.1 & 0.4 \\
\hline
Full model & 440.3 & 100.0 & 1210.0 & 100.0 \\
\hline
\end{tabular}
\caption{\textbf{Module-wise runtime breakdown of GeoUP.} \emph{Ratio} denotes the percentage of each module in the total module latency.}
\label{tab:appendix_runtime_breakdown}
\end{table}

To better understand the computational cost, we provide a module-wise runtime breakdown in Table~\ref{tab:appendix_runtime_breakdown}. 
For clarity, each task head includes its corresponding lightweight neck and prediction head. 
The auxiliary 2D branch has negligible measured latency in this profiling and is included in the detection head.
The runtime breakdown shows that the shared geometry backbone dominates the total latency. 
In the single-frame setting, the backbone takes 292.7 ms, accounting for 66.5\% of the total module latency. 
In the 4-frame setting, the backbone latency increases to 1062.1 ms and accounts for 87.8\% of the computation. 
In contrast, the task heads introduce relatively small and stable overheads. 
For example, the detection head remains around 28.6 ms, the depth head remains around 32.5 ms, and the occupancy head remains around 81.6 ms across the two settings. 

These results suggest that the most direct direction for improving GeoUP efficiency is to optimize the geometry-grounded backbone, adopt a lighter base model, and reduce the computational burden introduced by multi-frame processing. 
Another promising direction is to design a more unified decoder that shares computation across task readouts and processes multiple heads in parallel, instead of relying on separate task-specific decoders.

\section{Additional Quantitative Results}

\noindent\textbf{Additional low-resolution multi-dataset results.}
We further investigate the impact of multi-dataset training under low-resolution settings.
Specifically, we jointly train GeoUP on nuScenes~\cite{nuScenes}, Argoverse 2~\cite{Argoverse2}, Waymo~\cite{Waymo}, DDAD~\cite{DDAD}, and KITTI~\cite{KITTI}, where each dataset only supervises the heads with available annotations. 
For the input resolution, we use $672\times224$ for nuScenes~\cite{nuScenes}, $448\times336$ for Argoverse 2~\cite{Argoverse2}, and $504\times336$ for Waymo~\cite{Waymo}. 
As shown in Table~\ref{tab:appendix_multi_dataset}, multi-dataset training consistently improves the results.
Compared with single-dataset training, the jointly trained model achieves better detection and occupancy performance across different benchmarks. 
This demonstrates that heterogeneous supervision from datasets with different sensor layouts, scene distributions, perception ranges, and annotation types helps GeoUP learn a more robust geometry-grounded representation. 
These results further support the effectiveness of multi-dataset training for improving the generalization ability of unified 3D perception models.

\begin{table*}[t]
\centering
\scriptsize
\setlength{\tabcolsep}{9pt}
\renewcommand{\arraystretch}{1.2}

\begin{minipage}[t]{0.32\textwidth}
\centering
\textbf{(a) Argoverse 2}

\vspace{1mm}
\begin{tabular}{l|cc}
\hline
Setting & mAP$\uparrow$ & CDS$\uparrow$ \\
\hline
Single & 23.3 & 17.6 \\
Multi.  & \textbf{28.4} & \textbf{22.0} \\
\hline
\end{tabular}
\end{minipage}
\hspace{0.04\textwidth}
\begin{minipage}[t]{0.43\textwidth}
\centering
\textbf{(b) Waymo}

\vspace{1mm}
\begin{tabular}{l|ccc}
\hline
Setting & mAPL$\uparrow$ & mAP$\uparrow$ & mAPH$\uparrow$ \\
\hline
Single & 43.4 & 58.7 & 55.4 \\
Multi.  & \textbf{44.6} & \textbf{60.1} & \textbf{57.0} \\
\hline
\end{tabular}
\end{minipage}

\vspace{2mm}

\begin{minipage}[t]{0.62\textwidth}
\centering
\textbf{(c) nuScenes}

\vspace{1mm}
\begin{tabular}{l|cccc}
\hline
Setting & mAP$\uparrow$ & NDS$\uparrow$ & mIoU$\uparrow$ & RayIoU$\uparrow$ \\
\hline
Single & 56.5 & 63.6 & 41.6 & 46.1 \\
Multi.  & \textbf{57.6} & \textbf{64.1} & \textbf{42.3} & \textbf{47.0} \\
\hline
\end{tabular}
\end{minipage}

\caption{\textbf{Additional low-resolution results of multi-dataset training.} The joint model is trained on nuScenes~\cite{nuScenes}, Argoverse 2~\cite{Argoverse2}, Waymo~\cite{Waymo}, DDAD~\cite{DDAD}, and KITTI~\cite{KITTI}, where each dataset only supervises the heads with available annotations.}
\label{tab:appendix_multi_dataset}
\end{table*}

\noindent\textbf{Effect of camera pose supervision.}
\label{camera}
We further study the effect of the auxiliary camera pose supervision. 
Although the predicted camera pose is not directly used by the downstream perception heads during inference, it provides an explicit camera-level geometric constraint during training. 
As shown in Table~\ref{tab:appendix_camera_pose}, removing camera pose supervision consistently degrades detection and occupancy performance. 
This indicates that camera pose prediction is not merely an auxiliary output, but helps the shared backbone organize multi-view observations into a more metric and geometry-grounded 3D scene latent. 
Such a latent benefits both instance-level detection and volume-level occupancy prediction.

\begin{table}[t]
\centering
\small
\begin{tabular}{l|cccc}
\hline
Setting & mAP$\uparrow$ & NDS$\uparrow$ & mIoU$\uparrow$ & RayIoU$\uparrow$ \\
\hline
w/o camera pose & 55.3 & 62.4 & 40.1 & 43.9 \\
w/ camera pose & \textbf{56.4} & \textbf{62.9} & \textbf{40.6} & \textbf{44.1} \\
\hline
\end{tabular}

\caption{\textbf{Effect of camera pose supervision} on nuScenes~\cite{nuScenes}. The camera pose branch is only used as auxiliary supervision during training and is not directly used by downstream perception heads during inference.}
\label{tab:appendix_camera_pose}
\end{table}

\section{More Qualitative Results.}
\begin{figure}[t]
\centering
\includegraphics[width=0.9\linewidth]{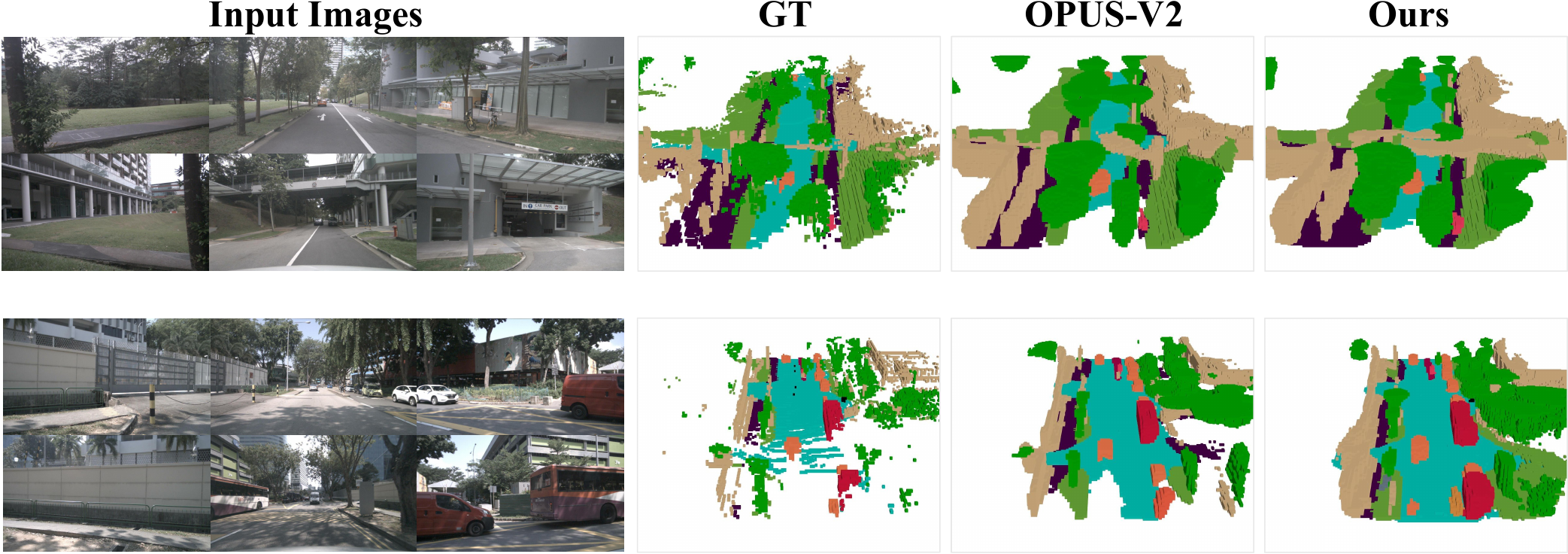}
\caption{
\textbf{Occupancy prediction} on Occ3D-nuScenes~\cite{occ3d}. GeoUP produces more complete occupancy predictions than OPUS-V2~\cite{OPUS}.
}
\label{fig:vis_occ_supp}
\end{figure}

\begin{figure}[t]
\centering
\includegraphics[width=0.9\linewidth] {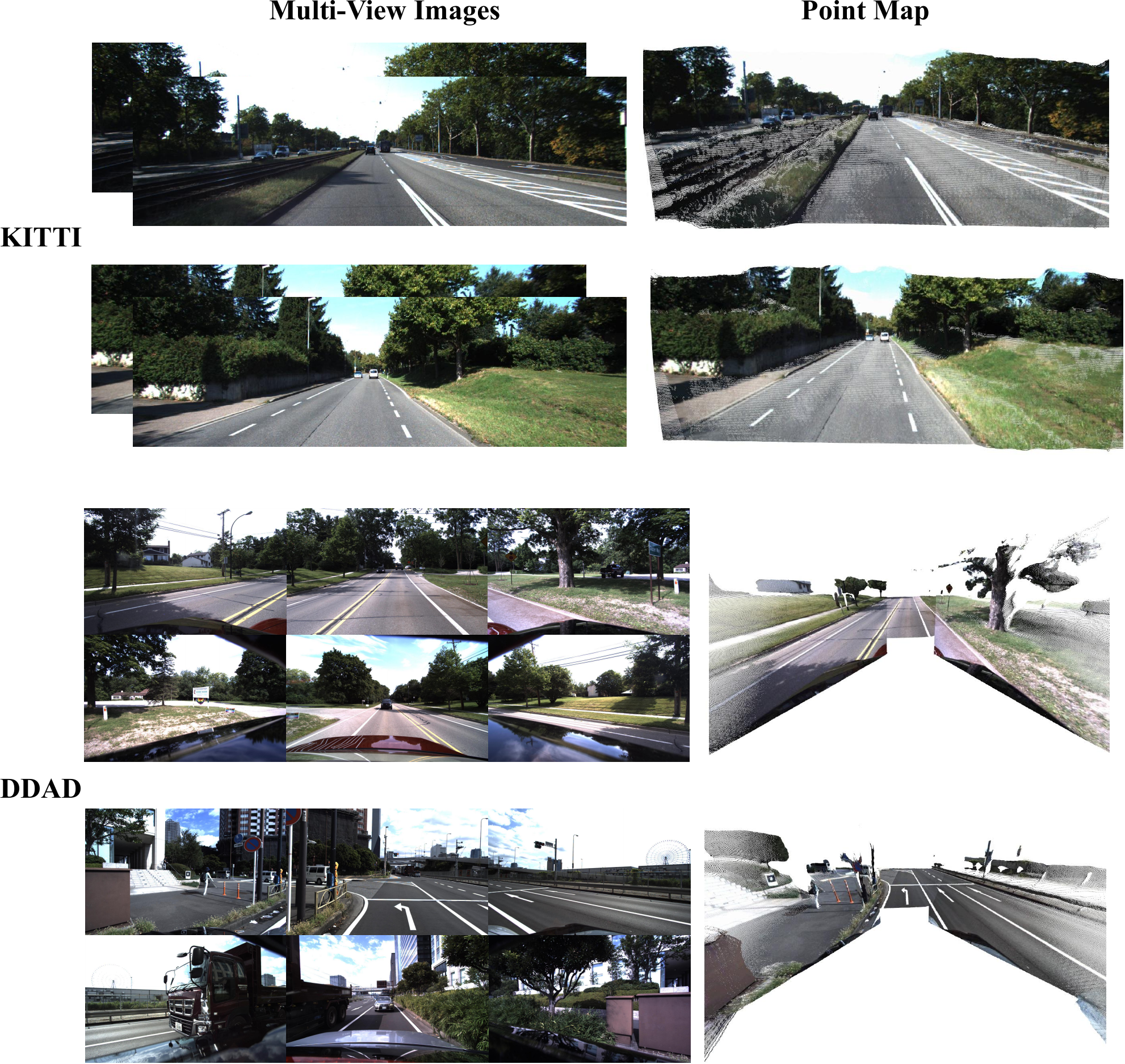}
\caption{
\textbf{Point maps} generated from predicted depth on KITTI~\cite{KITTI} and DDAD~\cite{DDAD}.
}
\label{fig:vis_depth_supp}
\end{figure}

\begin{figure}[t]
\centering
\includegraphics[width=0.9\linewidth]{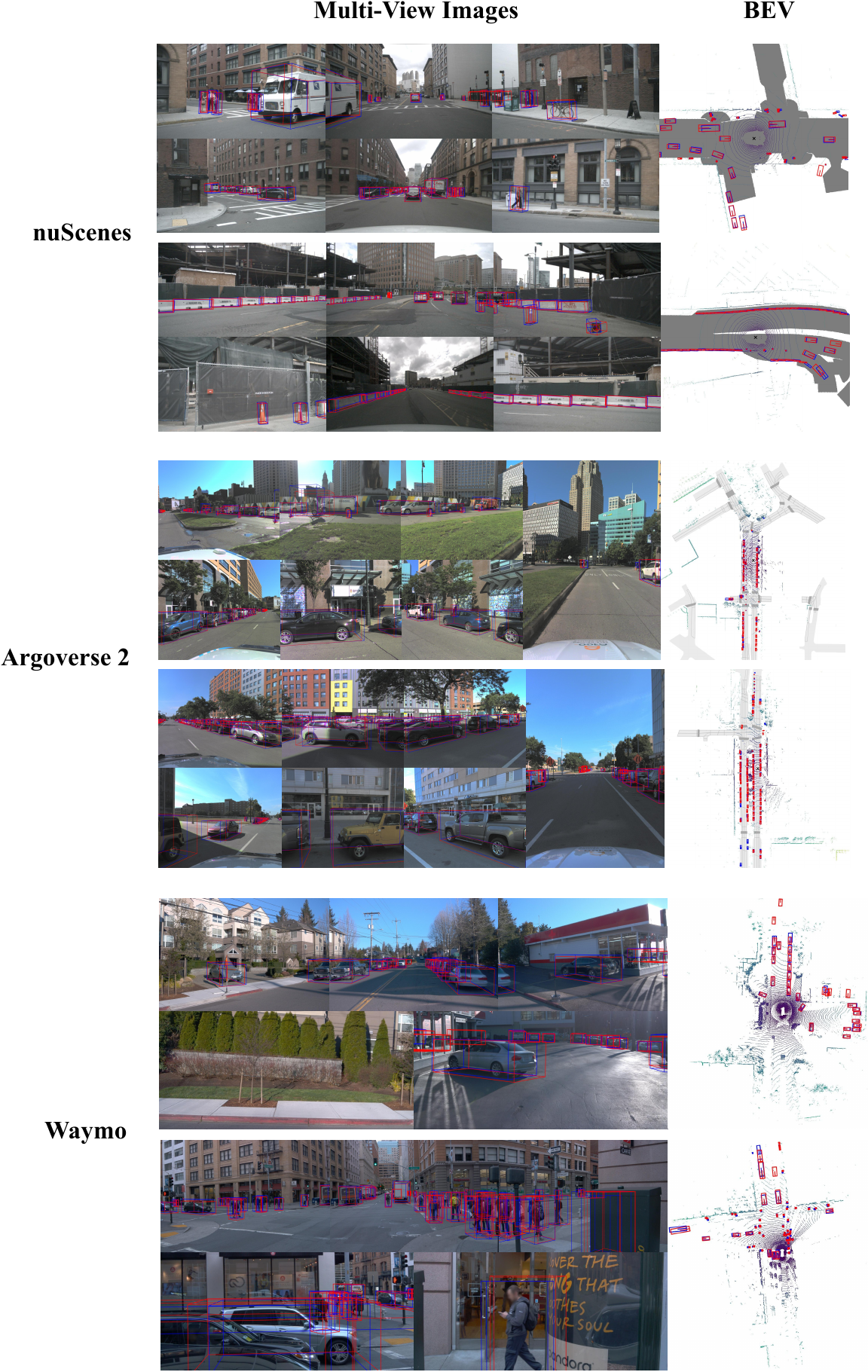}
\caption{
\textbf{3D object detection} on nuScenes~\cite{nuScenes}, Argoverse 2~\cite{Argoverse2}, and Waymo~\cite{Waymo}.
Blue boxes denote ground-truth annotations, and red boxes denote predictions.
}
\label{fig:vis_det_supp}
\end{figure}

We provide more qualitative results in Figures~\ref{fig:vis_occ_supp}, \ref{fig:vis_depth_supp}, and \ref{fig:vis_det_supp}. These visualizations cover different datasets and perception tasks, including semantic occupancy prediction on Occ3D-nuScenes~\cite{occ3d}, point map visualization generated from predicted depth on KITTI~\cite{KITTI} and DDAD~\cite{DDAD}, and 3D object detection on nuScenes~\cite{nuScenes}, Argoverse 2~\cite{Argoverse2}, and Waymo~\cite{Waymo}. The results further show that GeoUP produces reliable perception outputs across diverse sensor configurations, driving scenes, and task settings.

\end{document}